\documentclass{article} % For LaTeX2e
\usepackage[final]{colm2026_conference}

\usepackage{microtype}
\usepackage{hyperref}
\usepackage{url}
\usepackage{booktabs}

\usepackage{lineno}

\definecolor{darkblue}{rgb}{0, 0, 0.5}
\hypersetup{colorlinks=true, citecolor=darkblue, linkcolor=darkblue, urlcolor=darkblue}

\usepackage{graphicx}

\usepackage{tikz}
\usepackage{pgfplots}
\usepackage{colortbl}
\usepackage{xcolor}
\pgfplotsset{compat=1.17}

\usepackage{tcolorbox}
\usepackage{caption}
\usepackage{subcaption}

\definecolor{xy-axis}{HTML}{FFFDE5}
\definecolor{data-appearance}{HTML}{E5E5FF}
\definecolor{title-subtitle}{HTML}{E5E5E5}
\definecolor{legend}{HTML}{FFE5E5}
\definecolor{annotation}{HTML}{FBE7EB}
\definecolor{grid}{HTML}{F9F2EC}
\definecolor{figure-format}{HTML}{E2F3F8}
\definecolor{others}{HTML}{F2F2F2}

\definecolor{hu-made}{HTML}{3166FF}

\usepackage{booktabs}
\usepackage{listings}
\usepackage{colortbl}
\usepackage{makecell}

\usepackage{hyperref}
\usepackage{algorithm}
\usepackage{algpseudocode}
\usepackage{multirow}
\usepackage{amsmath}

\title{VisCodeBench: a new benchmark dataset for Text-to-Vis \\reflecting real-world practice}

\author{Hy Nguyen$^1$$^*$, Thomas James Elton$^1$, Ella Wilson$^1$, Andrew Reeson$^2$, Cécile Paris$^2$, \\ \textbf{Josiah Poon}$^1$, \textbf{Jonathan K. Kummerfeld}$^1$$^\dagger$ \\
The University of Sydney$^1$ \quad  CSIRO's Data61$^2$ \\
\texttt{nngu0448@uni.sydney.edu.au}$^*$ \quad \texttt{jonathan.kummerfeld@sydney.edu.au}$^\dagger$
}

\begin{document}

\ifcolmsubmission
\linenumbers
\fi

\maketitle

\begin{abstract}
Generating visualisations from natural language requests (text-to-vis) enables users to explore data more efficiently. 
While LLMs can complete simple visualisation requests, their performance on challenging ones is not well covered by existing benchmarks.
We introduce VisCodeBench, an end-to-end text-to-vis benchmark with 2,672 samples from three different domains:  educational, scientific, and general audience scenarios with varying levels of complexity. 
We benchmark three visualisation systems---Codex, Claude Code, and LIDA---paired with a range of open and commercial LLMs across seven metrics covering execution, chart content, and visual quality.
While basic execution is near-solved for strong systems, no system comes close to reproducing the reference data, with the best displayed data similarity reaching only $0.81$ and falling to $0.69$ on our hardest subset for this metric.
We further find that frontier models separate on chart correctness and efficiency, with ClaudeCode-Opus-4.7 improving chart correctness by around 5 points over Codex-GPT-5 while running roughly 20 seconds faster. 
Separately, we find that system design matters as much as the underlying model: despite both running on GPT-5, Codex clearly outperforms LIDA, showing that the surrounding prompting, execution, and refinement pipeline can shape outcomes as much as the model itself.
Finally, latency remains a practical obstacle, as even the strongest systems take 39-78 seconds to produce results.
VisCodeBench is available at \href{https://github.com/SAIPLL/VisCodeBench}{https://github.com/SAIPLL/VisCodeBench} and accessed via a leaderboard; the full dataset is kept private to prevent LLM training use.
\end{abstract}

\section{Introduction}
Text-to-Vis is the task of taking a human language request for visualisation of some data and generating code to produce the corresponding visualisation.
A range of benchmarks exist for this task, including nvBench~\citep{10.1145/3448016.3457261}, nvBench2.0~\citep{luo2025nvbench20resolvingambiguity}, ChartX~\citep{xia2024chartx}, 
Text2Chart31~\citep{pesaran-zadeh-etal-2024-text2chart31},
MatPlotBench~\citep{yang-etal-2024-matplotagent}, and
Text2Vis~\citep{rahman-etal-2025-text2vis}.
However, most of these datasets are synthesised or automatically constructed and do not match real-world use, as shown in our earlier work \citep{nguyen-etal-2024-text}.
Often, the visualisations lack meaningful real-world context, or the generated code follows fixed templates.

We introduce VisCodeBench, a new benchmark dataset for evaluating text-to-vis systems in real-world scenarios. 
VisCodeBench is based on three sources of data.
(1) GitHub-T2V contains 843 samples based on assignment notebooks used in online courses shared on GitHub.
(2) arXiv-T2V is comprised of 805 samples from academic papers, collected by contacting authors and requesting the chart, code, underlying data, and a natural language request describing the visualisation.
(3) OWID-T2V includes 1,024 samples from \href{https://ourworldindata.org}{https://ourworldindata.org}, a publicly accessible platform with data-driven analyses of global issues.
After collecting the raw data, expert annotators, each with over four years of coding experience and extensive experience creating academic visualisations, verified and refined the samples. 
A range of statistics show that the final datasets are more difficult and realistic than existing benchmarks.

For evaluation, we propose two new metrics in addition to five existing ones.
(1) Displayed Data Similarity assesses the similarity of data points displayed in generated and reference charts~\citep{8978105}, providing a more general, LLM-independent, and flexible alternative to exact-match evaluations used in existing work~\citep{10670425, yang-etal-2024-matplotagent};
(2) Accessibility Score measures contrast and spacing of texts displayed on the visualisation images~\citep{miranda-pena-etal-2024-llms}.

We benchmark three visualisation systems---Codex, Claude Code, and LIDA---paired with open (GPT-OSS-20B, Gemma-4-31B, Qwen-3.6-35B) and commercial (GPT-5, Opus-4.7) LLMs.
Our results show that basic execution is near-solved for strong systems, with only the small local model failing to reliably produce valid charts.
Content fidelity is the harder challenge, as no system reproduces the reference data well.
The best displayed data similarity comes from Claude Code with Opus-4.7, but it reaches only $0.81$ and falls to $0.69$ on OWID-T2V, the hardest subset for displayed data similarity.
VisCodeBench separates frontier models: they differ on chart correctness score and efficiency, with ClaudeCode-Opus-4.7 improving chart correctness by around 5 points over Codex-GPT-5 while running roughly 20 seconds faster, making Opus-4.7 the strongest model overall.
We further show that a system's design can weigh as heavily as its underlying model: Codex and LIDA share the same GPT-5 backbone, yet on OWID-T2V, LIDA completes only 79\% of tasks compared to Codex's 100\%, with correctness falling to 54.37\% versus 73.67\%.
Finally, latency remains a barrier to interactive use across every system we test.
Together, these findings highlight the challenges of realistic text-to-vis scenarios and demonstrate the value of VisCodeBench in exposing them.\footnote{The evaluation code, representative samples, leaderboard, and are available at \href{https://github.com/SAIPLL/VisCodeBench}{VisCodeBench GitHub}, and the full benchmark leaderboard is linked from that page.}

\begin{figure*}[t]
\centering
\includegraphics[width=0.9\textwidth]{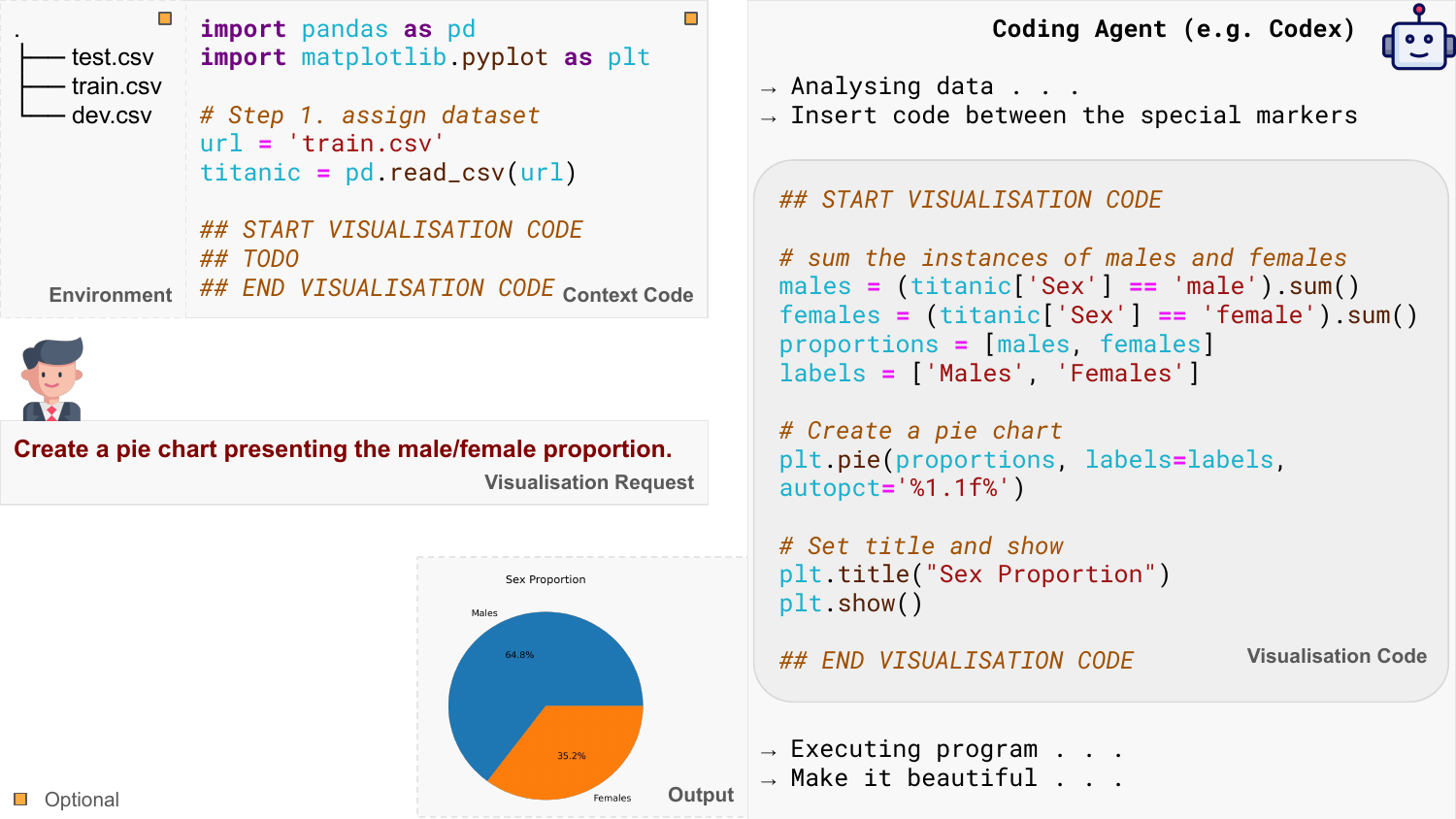}
\caption{A Text-to-Vis system. Left: input data files, code context, and user request. Right: system output.}
\label{fig:text-to-vis-sys}
\end{figure*}

\section{Related Work}
\subsection{Development of Text-to-Vis Benchmarks}
Early Text-to-Vis benchmarks were manually created. 
For example, for the NLV Corpus~\citep{10.1145/3411764.3445400} and ChartDialogs~\citep{shao-nakashole-2020-chartdialogs}, crowd workers were given real-world data tables and charts. 
They then wrote commands that described how a system should create a chart, change its layout, or modify its visual style. 
This method is closely aligned with real-world use because the charts, data, and requests originate directly from people.
However, it has clear limits: only a few chart types are used, the data tables are small, and the natural language requests are in a single turn or short. 

To address these limits, researchers moved to semi-automated and later fully automated approaches. 
Semi-automated datasets—such as nvBench~\citep{10.1145/3448016.3457261}, 
Dial-nvBench~\citep{song2024marrying}, VisEval~\citep{10670425}, 
MatplotBench~\citep{yang-etal-2024-matplotagent},
Text2Vis~\citep{rahman-etal-2025-text2vis},
and nvBench2.0~\citep{luo2025nvbench20resolvingambiguity}
—begin with a real-world data table, from which a synthesiser generates a chart and a corresponding natural language request; then human annotators verify and refine the results.
Fully automated approaches take this further by using LLMs to generate the data table, natural language request, code, and visualisation from pre-designed prompts, e.g., for Chartllama~\citep{han2023chartllama}, ChartX~\citep{xia2024chartx}, and Text2Chart31~\citep{pesaran-zadeh-etal-2024-text2chart31}.
Although these methods allow rapid dataset scaling and control over diversity and balance, they tend to be less realistic: the data is often synthetic, the charts convey little meaningful real-world insight, and the outputs lack the customisation typical of genuine use.
Also, the requests and visualisations often follow strict templates made by dataset designers, making them less like real user requests.

In our work, we source charts from data created by real users, avoiding the issues with the synthetic worker-written data raised above.
To ensure quality, we then use human effort with AI assistance to refine code and requests.

\subsection{Automated Evaluation for Text-to-Vis}
% For instance, the reference command \texttt{plt.scatter(age, duration)} specifies a scatter plot with \texttt{age} on the x-axis and \texttt{duration} on the y-axis. If the generated program produces the same function call with the same variables, it is deemed correct.
% While this approach works well for structured visualisation languages such as VQL and Vega-Lite, it is highly language-dependent and difficult to generalise. 
% In more complex programming languages like Python, it requires substantial effort to normalise code and perform strict comparisons. For example, \texttt{plt.scatter(age, dur)} should still be considered correct when \texttt{dur = duration}, yet such equivalences are non-trivial to detect automatically.

\textbf{Execution-based Evaluation Methods} evaluate visualisations by executing the generated programs and comparing the resulting charts with reference chart outputs in terms of chart type and displayed data.
To compare the displayed data, prior work commonly uses the extracted x- and y-coordinates, obtained either by parsing the rendered output (e.g., from SVG files) or by logging values directly during execution~\citep{10121440, li2024visualization, 10670425}.
While this approach directly verifies whether a visualisation encodes the intended data, it introduces several challenges. 
First, it requires substantial post-processing, such as normalising numerical values and enforcing consistent naming for temporal units. 
Second, the comparison is often overly strict: a single misplaced or incorrectly scaled data point can cause the entire output to be marked as incorrect.

\textbf{LLM-based Evaluation Methods} employ LLMs (e.g., GPT-4o) to assess the readability and quality of generated chart images, or their correctness against reference charts~\citep{dibia-2023-lida, 10670425, rahman-etal-2025-text2vis, yang-etal-2024-matplotagent}. 
This approach is easy to implement and correlates well with human judgment. 
However, it depends on proprietary models and incurs substantial cost when scaled to large datasets.

To address these limitations, we adopt Displayed Data Similarity, a more flexible metric that measures the similarity between the two sets of displayed data points. Inspired by CHART-Infographics~\citep{8978105}, a competition focused on extracting raw data from pixel-image charts, our metric offers a more tolerant, informative, and LLM-independent measure of performance.

\section{Dataset Construction}
\label{sec:dataset-construction}
Our benchmark includes 2,672 samples, grouped into three subsets: \textbf{GitHub-T2V}, \textbf{arXiv-T2V}, and \textbf{OWID-T2V}.
After collection, the raw data went through an annotation process involving both LLM-assisted and human interventions, followed by expert manual review.
The dataset is formatted as illustrated in Figure~\ref{fig:text-to-vis-sys} to facilitate easy benchmarking of coding agents, such as Codex and Claude Code.
% Each sample includes inputs comprising environment files, contextual code, and a natural language visualisation request, along with outputs consisting of the reference visualisation code and the corresponding rendered chart.

\subsection{GitHub-T2V}
\label{ssec:github-t2v}
GitHub-T2V contains 843 text-to-vis samples collected from an educational context, where students write visualisation code as part of assignments in data science, data analysis, and machine learning courses. 
Each sample includes the underlying data and dependent files, the relevant context code, a natural language visualisation request, the corresponding visualisation code, and the resulting chart.

The construction process followed a five-step pipeline. First, we scanned The Stack~\footnote{\href{https://huggingface.co/datasets/bigcode/the-stack}{https://huggingface.co/datasets/bigcode/the-stack}}, a large-scale pre-training dataset for code LLMs, to identify GitHub URLs of MIT-licensed Python notebooks whose filenames contain ``solution'', indicating homework solutions or teaching materials. 
Second, we randomly sampled 5,000 notebooks and retrieved their associated environment files, including imported modules and underlying data files, from the original GitHub repositories. 
Third, we automatically executed the notebooks using a custom script, retaining only those that ran successfully within a 60-second time limit and discarding notebooks with runtime errors.
Fourth, we used GPT-4o to automatically detect natural-language visualisation requests, which typically appear in markdown cells immediately preceding code cells that generate charts. 
Finally, we manually verified the detected natural language visualisation requests and constructed samples in our format.

\subsection{arXiv-T2V}
\label{ssec:arxiv-t2v}
arXiv-T2V is comprised of 805 text-to-vis samples collected from scientific research settings, where researchers write code to visualise experimental data using graphs and charts.
The raw data for this subset were collected via an online survey approved by our organisations' ethics committees.\footnote{For details of ethics approval, see the Ethics Statement.}
We invited approximately 30,000 authors of CC BY-SA 4.0–licensed papers published on arXiv in 2024 using the email addresses listed in their papers, and further promoted the study through social media and academic mailing lists.
Participants were asked to submit chart images from their papers, along with the corresponding code, underlying data, and a newly written natural language request describing how they would ask another person to create the visualisation.
In total, we received 1,151 valid submissions from 848 participants, corresponding to a response rate of 2.69\%.
Additional details are provided in Appendix~\ref{sec:online-survey-arXiv}.

% We release the original participant-provided requests alongside the dataset as supplementary reference material.

We filtered down to suitable visualisations and modified the code to suit our benchmark.
We manually de-identified and selected 805 submissions that target Python-based visualisation code and involve substantial data processing and manipulation. 
We then determined the placement of the natural language request within the code: when the underlying data are provided as files, the request is inserted at the beginning of the Python code file; when the data are embedded directly in the code, the request is placed right after the data definition. 

Although we asked participants to write a natural language request, many of the submitted requests were overly abstract or appeared to be generated by large language models.
To address this issue, we created new natural language requests for the charts.
We used GPT-5 to generate initial natural language requests of approximately 120 words, ensuring that the requests were concrete and sufficiently detailed.
Native English–speaking experts with over four years of programming experience in Python-based academic data visualisation revised the initial requests. 
During this stage, the experts were provided only with the raw data (data files or code-embedded data), the initial requests, and the resulting visualisation images-they were not given access to the visualisation code. 
They were advised to explore and visualise the data by writing and executing their own code until they developed a clear understanding of the data and the visualisations.
The experts then revised the initial requests, phrasing them as they would when prompting a coding agent to generate the visualisations from the given data.

\begin{figure}[t!]
\centering
\includegraphics[width=0.4\textwidth]{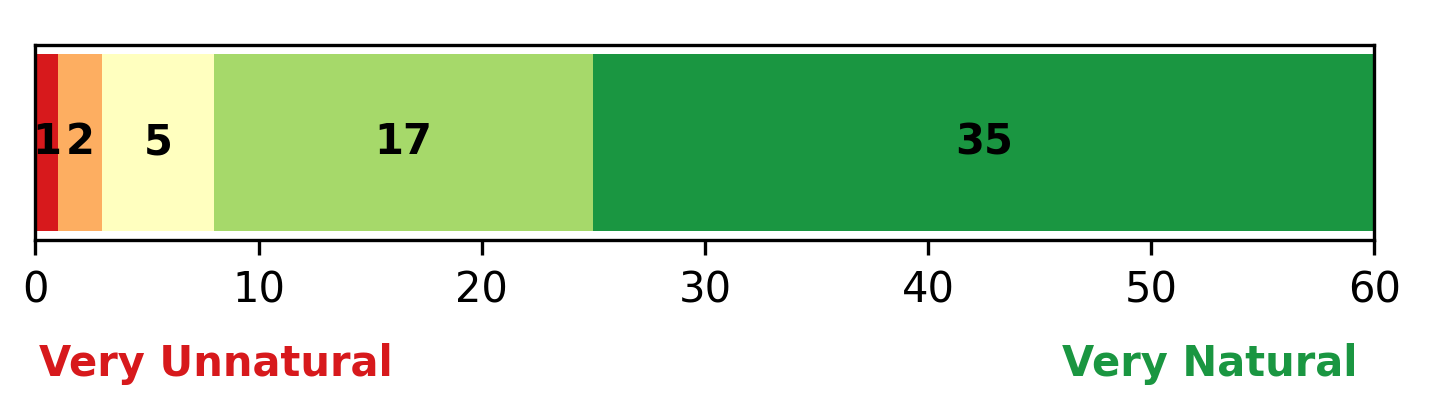}
\caption{Annotation quality assessment of arXiv-T2V}
\label{fig:assess-annotation-quality}
\end{figure}

\textbf{Annotation Quality}\quad
To assess annotation quality, an expert with over ten years of programming experience and extensive involvement in academic data visualisation manually rated 60 samples.
The expert had full access to each annotated sample, 
%including the underlying data, the natural language request, the visualisation code, and the resulting output image, 
and was asked to rate the naturalness of the request on a five-point Likert scale (1 = very unnatural, 5 = very natural).
As shown in Figure~\ref{fig:assess-annotation-quality}, the score distribution shows that most samples received high ratings: 52 out of 60 requests received scores of 4 or 5, with an average score of 4.34, indicating a generally high level of naturalness.

The expert was also asked to provide reasons for any rating below 5. 
The most common issue was the use of excessively long paragraphs that combine multiple categories of information—such as data loading, data processing, the core visualisation request, and chart styling instructions. 
Other issues include the use of specialised mathematical symbols that are difficult to type on a standard keyboard, as well as over-specification, such as repeating items already present in the context or providing unnecessary details about numerical formats (e.g., decimals and percentages).

\subsection{OWID-T2V}
\label{ssec:owid-t2v}

OWID-T2V consists of 1,024 text-to-vis samples collected from \href{https://ourworldindata.org/}{https://ourworldindata.org/}, a publicly accessible research platform where data scientists and researchers curate datasets, develop visualisations, and write analytic articles on major global issues.
Each sample includes a underlying data file, a natural language visualisation request, and the corresponding chart image.
Since the original image was produced by custom javascript, we do not have code for these samples.
We instead provide the reference displayed data points and the original chart images for evaluation.
% and do not apply code-based evaluation metrics.
% This is because the original charts are generated by an application that involves both front-end and back-end components rather than a standalone Python script; therefore, we provide reference data points to enable quantitative evaluation of text-to-vis systems.

We built the data by collecting analytical articles from \href{https://ourworldindata.org/}{https://ourworldindata.org/}. 
For each article, we extracted the associated charts and related materials, including: (1) chart images; (2) the analytical discussion written by the article authors, in which the charts are used as supporting evidence; (3) chart- and dataset-level metadata; (4) the full CSV dataset used to generate each chart; and (5) the displayed CSV data shown in the chart after filtering and processing by the website.

To generate initial natural language requests, we used GPT-5.
The model was guided to write a 50-word request, following the wording and grammatical structure suggested by the chart and its associated materials, with particular emphasis on the analytical discussion in which the chart is referenced (e.g., ``The bar chart illustrates global births in 2025''). 
Expert annotators then revised these drafts to produce the final human-authored requests, as described in section~\ref{ssec:arxiv-t2v}.

Finally, we reconstructed the reference displayed data points in x-, y-, and z-coordinates from the published CSV data. 
The reference data points were created by: (1) using Codex with GPT-5 to replicate each chart image based on the published CSV data and metadata; (2) manually comparing the replicated charts with the original images and repeating this process until a visually similar result was obtained, discarding samples that fail to converge after three attempts; and (3) extracting the reference data points using custom scripts, which are described in Appendix~\ref{sec:appendix-data-extraction-metric}.

\section{Dataset Analysis}
\subsection{Statistics}
\begin{figure}[htbp]
    \centering
    \begin{subfigure}[b]{0.49\textwidth}
        \centering
        \includegraphics[width=0.8\textwidth]{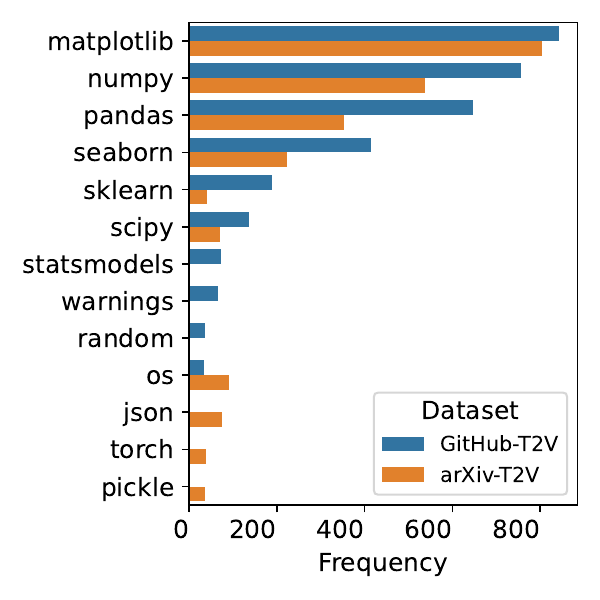}
        % \caption{Top 10 libraries used in VisCodeBench}
        % \label{fig:analysis-top-libraries-and-plot-types}
    \end{subfigure}
    \hfill
    \begin{subfigure}[b]{0.49\textwidth}
        \centering
        \includegraphics[width=0.8\textwidth]{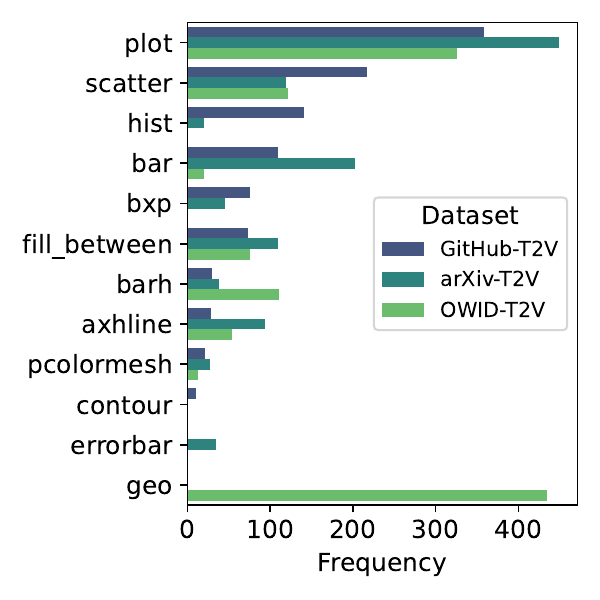}
        % \caption{Frequency of plot types in VisCodeBench}
        % \label{fig:analysis-plot-types}
    \end{subfigure}
    \caption{Top 10 libraries and plot types used in VisCodeBench.}
    \label{fig:analysis-top-libraries-and-plot-types}
\end{figure}

\textbf{Libraries Used}\quad
Figure~\ref{fig:analysis-top-libraries-and-plot-types} (left) shows the ten most commonly used Python libraries in GitHub-T2V and arXiv-T2V subsets.
Among them, \texttt{matplotlib}, \texttt{numpy}, \texttt{pandas}, \texttt{seaborn}, \texttt{sklearn}, and \texttt{scipy} stand out as the most prevalent, reflecting their essential roles in data manipulation, transformation, and visualisation.
Libraries such as \texttt{random}, \texttt{os}, \texttt{json}, and \texttt{pickle} also appear frequently, primarily for tasks like loading data, controlling randomness, and handling I/O.
The popularity of \texttt{statsmodels} on GitHub-T2V versus \texttt{torch} on arXiv reflects the distinct characteristics of these two datasets.

% \begin{figure}[t!]
% \centering
% \includegraphics[width=\columnwidth]{images/analysis-top-libraries.pdf}
% \caption{Top 10 libraries used in VisCodeBench}
% \label{fig:analysis-top-libraries}
% \end{figure}

\textbf{Plot type}\quad
Figure~\ref{fig:analysis-top-libraries-and-plot-types} (right) illustrates the distribution of plot types across our data subsets.
GitHub-T2V and arXiv-T2V exhibit very similar chart-type distributions: line charts are the most popular, followed by scatter plots, histograms, bar charts, box plots, and area charts.
The most striking outlier is the geomap category, which appears almost exclusively in OWID-T2V, reflecting the dataset’s emphasis on geographical data.

% \begin{figure}[t!]
% \centering
% \includegraphics[width=0.7\columnwidth]{images/analysis-plottypes.pdf}
% \caption{Frequency of plot types in VisCodeBench}
% \label{fig:analysis-plot-types}
% \end{figure}

\textbf{Analysis of other dimensions}\quad
Table~\ref{tab:analysis-comparison-with-related-benchmarks} (lower) presents statistics across various dimensions.
Overall, the arXiv-T2V dataset exhibits longer outputs, a greater variety of plot types, and more subplots, indicating a stronger need for customisation in both style and layout of the resulting visualisations. 
% This highlights the differences in data sources between the three subsets: GitHub-T2V is derived from online notebooks focused on data science and machine learning education, OWID-T2V is sourced from analytic articles on major global issues; whereas arXiv-T2V reflects visualisation use cases found in academic publications.

\subsection{Comparison with real-world code}
\label{ssec:comparison-with-real-world-code}
Following the approach from our earlier work~\citet{nguyen-etal-2024-text}, we compare code from our dataset subsets with real-world visualisation code collected from the Internet, focusing specifically on Matplotlib.
In terms of chart type distribution, GitHub-T2V and arXiv-T2V exhibit strong alignment with real-world data, with line graphs being the most common chart type, followed by scatter and bar charts.
For usage patterns of 62 common attributes, GitHub-T2V and arXiv-T2V show strong alignment with real-world data, with Spearman’s rank correlation coefficients exceeding 0.91.
Regarding complexity, while GitHub-T2V exhibits the same level of complexity as real-world data, arXiv-T2V is more complex, with an average of 9.20 distinct functions and 17.32 parameters per program.
Further details of these comparisons are provided in Appendix~\ref{sec:appendix-analysis2-realworld-comparison}.

\subsection{Comparison with related datasets}
We focus comparisons on seven public end-to-end benchmark datasets: NLV Corpus~\citep{10.1145/3411764.3445400}, nvBench~\citep{10.1145/3448016.3457261}, nvBench2.0~\citep{luo2025nvbench20resolvingambiguity}, ChartX~\citep{xia2024chartx}, Text2Chart31~\citep{pesaran-zadeh-etal-2024-text2chart31}, MatplotBench~\citep{yang-etal-2024-matplotagent}, and Text2Vis~\citep{rahman-etal-2025-text2vis}, as summarised in Table~\ref{tab:analysis-comparison-with-related-benchmarks}.
Other existing datasets are excluded because they are variants of these benchmarks or do not support end-to-end text-to-vis evaluation.

\begin{table*}[!t]
\centering
\small
\setlength{\tabcolsep}{2pt}
\resizebox{\textwidth}{!}{
\begin{tabular}{ccccccccc|cccc}
\toprule
 & Annotation & Test & Data & Plot & Distinct & Sub- & Input & Output & \multicolumn{2}{c}{Diversity} & \multicolumn{2}{c}{Complexity} \\
 & note & Samples & Format &Types & libs* & -plots* & Length* & Length* & Input & Output & \#func* & \#args* \\
\midrule
NLV Corpus & \textit{\textcolor{hu-made}{D}}, \textit{\textcolor{hu-made}{R}}, \textit{\textcolor{hu-made}{C}}, \textit{\textcolor{hu-made}{I}} & 893 & Fixed & 4 & 1.0 & 1.55 & 57.7 & 376 & \textbf{0.64} & 0.01 & 7.11 & 12.56 \\
nvBench & \textit{\textcolor{hu-made}{D}}, R, C, I & \textbf{3990} & Fixed & 4 & 1.0 & 1.00 & 131 & 709 & 0.13 & 0.01 & 4.59 & 10.02 \\
nvBench2.0 & \textit{\textcolor{hu-made}{D}}, R, C, I & 751 & Fixed & 6 & 1.0 & 1.00 & 3532 & 458 & 0.07 & 0.03 & 4.94 & 5.54 \\
ChartX & D, R, C, I & 1152 & Fixed & 16 & 1.98 & 1.18 & 1030 & 1159 & 0.20 & 0.09 & 10.19 & 16.03 \\
Text2Chart31 & D, R, C, I & 1423 & Fixed & \textbf{31} & 2.46 & 1.08 & 1069 & 767 & 0.38 & 0.18 & 8.64 & 11.09 \\
MatPlotBench & \textit{\textcolor{hu-made}{D}}, \textit{\textcolor{hu-made}{R}}, \textit{\textcolor{hu-made}{I}} & 100 & Fixed & 13 & - & 1.55 & 1532  & - & 0.62 & - & - & - \\
% ChartMimic & \textit{\textcolor{hu-made}{D}}, \textit{\textcolor{hu-made}{R}}, \textit{\textcolor{hu-made}{C}}, \textit{\textcolor{hu-made}{I}} & \textbf{4800} & 22 & 2.19 & 1.61  & 2127 & 2363 & 0.08 & 0.06 &  \textbf{11.3} & \textbf{23.1} \\
Text2Vis & \textit{\textcolor{hu-made}{D}}, \textit{\textcolor{hu-made}{R}}, C, I & 1985 & Fixed & 20 & 1.71 & 1.15 & 1153 & 1219 & 0.18 & 0.16 & 9.02 &  14.2\\
\midrule
\textbf{GitHub-T2V} & \textit{\textcolor{hu-made}{D}}, \textit{\textcolor{hu-made}{R}}, \textit{\textcolor{hu-made}{C}}, \textit{\textcolor{hu-made}{I}} & 843 & \textbf{Free} & 18 & \textbf{4.08} & 1.96 & \textbf{6745} & 227 & 0.13 & \textbf{0.31} & 10.34 & 15.61 \\
\textbf{arXiv-T2V} & \textit{\textcolor{hu-made}{D}}, \textit{\textcolor{hu-made}{R}}, \textit{\textcolor{hu-made}{C}}, \textit{\textcolor{hu-made}{I}} & 805 & \textbf{Free} & 21 & 3.15 & \textbf{2.54} & 1557 & \textbf{2847} & 0.46 & 0.20 & \textbf{10.54} & \textbf{20.37} \\
\textbf{OWID-T2V} & \textit{\textcolor{hu-made}{D}}, \textit{\textcolor{hu-made}{R}}, \textit{\textcolor{hu-made}{DD}}, \textit{\textcolor{hu-made}{I}} & 1024 & Fixed & 6 & - & 1.15 & 209 & - & 0.55 & - & - & - \\
\bottomrule
\end{tabular}
}
\caption{Comparison with related benchmarks.
Metrics with a * are averaged over samples.
D, R, C, and I indicate underlying data, natural language request, code, and chart image, respectively. Annotation notes displayed in italic blue text denote components that were created by \textit{\textcolor{hu-made}{humans}}. 
Note, OWID-T2V does not have code, which is why Displayed Data (\textit{\textcolor{hu-made}{DD}}) is provided for evaluation and some values are blank. 
Data Format denotes whether the underlying data has a \textbf{fixed} structure (e.g., CSV) or varies \textbf{free}ly in schema and format.}
\label{tab:analysis-comparison-with-related-benchmarks}
\end{table*}

Compared to existing datasets, ours show significant advantages in input/output length, the number of distinct libraries used per sample, the presence of multiple subplots per chart, and the flexibility of data format.
These characteristics indicate that our benchmark dataset places higher demands on models, requiring stronger contextual understanding, the ability to generate longer and more complex code, and the capacity to produce charts with multiple subplots and customisations. 
In comparison to the NLV Corpus, a fully human-curated dataset, our dataset is comparable in test set size but supports a wider range of plot types.

We also measured content diversity for both inputs and outputs, tokenised with the GPT-2 tokenizer~\citep{radford2019language}.
arXiv-T2V achieves high diversity scores for both inputs and outputs, despite processing longer input sequences and generating substantially longer code outputs than the other datasets. 
This result is noteworthy because diversity measured by distinct n-grams typically decreases with increasing sequence length due to repetition~\citep{li-etal-2016-diversity}.

We further evaluated the complexity of visualisation programs using the metric we proposed in prior work~\citet{nguyen-etal-2024-text}.
In this analysis, we considered visualisation functions from Matplotlib, Pandas, and Seaborn, rather than limiting the evaluation to Matplotlib alone, as computed in section~\ref{ssec:comparison-with-real-world-code}.
As shown in the last two columns of Table~\ref{tab:analysis-comparison-with-related-benchmarks}, our arXiv-T2V exhibits high complexity, with an average of approximately 10.54 distinct visualisation functions and 20.37 visualisation parameters per program.

\section{Evaluation Metrics}
We implemented seven evaluation metrics, which differ in their evaluation targets and level of granularity. 
Together, they provide a comprehensive assessment of system performance in terms of both data presented and presentation style.

\textbf{Task Completion Rate (TCR)}\quad
We compute $TCR = \frac{C}{T}$. 
Where $C$ is how many samples the generated code runs successfully on and produces at least one visualisation image, and $T$ is the total number of samples.

\textbf{Prediction Time}\quad
This is the time spent, measuring the total time required to complete the visualisation task.

\textbf{Chart Type Accuracy}\quad
This measures whether the predicted and reference chart types match.
In our benchmark, we implement this metric to consider multiple chart types in both generated and reference charts.
Let $P=\{p_1,\ldots,p_m\}$ and $G=\{g_1,\ldots,g_n\}$ denote the set of chart types in the generated and reference visualisations. 
A prediction is considered correct if $|P \cap G| = |G|$.

\textbf{Displayed Data Similarity}\quad
After extracting displayed data points using a monkey-patching technique (see Appendix~\ref{sec:appendix-data-extraction-metric} for details), we adopt the evaluation metric from Task~6b of the ICDAR 2019 Competition on Harvesting Raw Tables from Infographics~\citep{8978105}, with several modifications to better suit our setting.
(1) When comparing data series from two lines, we replace linear interpolation, which causes errors for area charts containing cycles, with arc-length–parameterised interpolation to establish stable pointwise correspondence along the curves~\citep{farin2014curves}.
(2) We extend the original data-series–level similarity computation to figures with multiple subplots: subplot scores are computed from their constituent data series, and the overall figure score is obtained by aggregating the subplot scores.

\textbf{Chart Correctness}\quad
We adopt the chart correctness evaluation method proposed by~\citet{yang-etal-2024-matplotagent}, which employs an LLM as a judge to assign a correctness score on a 100-point scale to each pair of predicted and reference chart images.

\textbf{Readability Score}\quad
We adopt the readability evaluation method proposed by~\citet{10670425}, which employs an LLM as a judge to assign a readability score on a five-point scale to each predicted chart image. 

\textbf{Accessibility Scores}\quad
We follow~\citet{miranda-pena-etal-2024-llms} with two accessibility scores: text colour contrast and text spacing. 
These metrics evaluate the readability of chart text by measuring the contrast between text and background colours, as well as character spacing.

\section{Benchmarking Visualisation Agents}
\label{sec:benchmarking-coding-agents}

In this work, we used our dataset to benchmark three visualisation systems: two general-purpose LLM-powered coding agents, Codex (Co)\footnote{\href{https://github.com/openai/codex}{https://github.com/openai/codex}} and Claude Code (Cl)\footnote{\href{https://github.com/anthropics/claude-code}{https://github.com/anthropics/claude-code}}, and one dedicated text-to-vis pipeline that constructs the visualisation through separate components, LIDA~\citep{dibia-2023-lida}, 
For the underlying LLMs, we selected models that reflect common deployment scenarios.
The commercial models GPT-5 and Opus-4.7 process data externally with fewer hardware constraints, whereas the open models GPT-OSS-20B, Gemma-4-31B, and Qwen-3.6-35B run locally for privacy-sensitive or resource-limited deployments.
Additional details about the experimental setup are provided in Appendix~\ref{sec:appendix-setting-up-coding-agents}.
The overall performance of the systems is summarised in Table~\ref{tab:performance-of-coding-agents}.
% For each sample in each subset, we ran each coding agent three times to generate three candidate programs and then take the average metric scores across the three runs. 
% We then applied the evaluation metrics to these generated programs and reported the final results as the average scores 

\begin{table*}[t]
\centering
\footnotesize
\setlength{\tabcolsep}{3pt}
\begin{tabular}{llrrrrrrrr}
\toprule
\textbf{Subset} & \textbf{Model} & \textbf{Task} & \textbf{Pred.} & \textbf{Chart} & \textbf{Data} & \textbf{Correct-} & \textbf{Reada-} & \textbf{Text} & \textbf{Text} \\
 &  & \textbf{Compl.} & \textbf{Time} & \textbf{Type} & \textbf{Sim.} & \textbf{ness} & \textbf{bility} & \textbf{Contrast} & \textbf{Spacing} \\
\midrule
\multirow{6}{*}{GitHub-T2V} & Co-GPT-OSS-20B & 0.76 & 66s & 0.64 & 0.49 & 51.71 & 4.80 & 0.90 & $-$0.42 \\
 & Cl-Qwen3.6-35B & 0.96 & 93s & 0.86 & 0.60 & 68.06 & 4.81 & 0.88 & $-$2.12 \\
 & CL-Gemma4-31B & 0.99 & 90s & 0.90 & \textbf{0.70} & \textbf{75.41} & 4.83 & 0.88 & $-$0.33 \\
 & LIDA-GPT-5 & 0.98 & 113s & 0.90 & 0.66 & 71.74 & 4.71 & 0.89 & $-$0.58 \\
 & Co-GPT-5 & \textbf{1.00} & 68s & 0.90 & 0.61 & 69.92 & 4.85 & 0.88 & $-$0.98 \\
 & Cl-Opus-4.7 & \textbf{1.00} & \textbf{39s} & \textbf{0.92} & 0.68 & 74.68 & 4.91 & 0.88 & $-$2.20 \\
\midrule
\multirow{6}{*}{arXiv-T2V} & Co-GPT-OSS-20B & 0.67 & \textbf{55s} & 0.62 & 0.48 & 42.20 & 4.54 & 0.89 & $-$0.96 \\
 & Cl-Qwen3.6-35B & 0.97 & 231s & 0.92 & 0.73 & 68.07 & 4.69 & 0.88 & $-$3.51 \\
 & CL-Gemma4-31B & 0.99 & 124s & \textbf{0.94} & 0.78 & 70.80 & 4.72 & 0.86 & $-$1.35 \\
 & LIDA-GPT-5 & 0.94 & 124s & 0.88 & 0.74 & 68.03 & 4.55 & 0.88 & $-$1.43 \\
 & Co-GPT-5 & \textbf{1.00} & 75s & 0.93 & 0.80 & 72.36 & 4.72 & 0.88 & $-$1.98 \\
 & Cl-Opus-4.7 & \textbf{1.00} & \textbf{55s} & 0.93 & \textbf{0.81} & \textbf{77.03} & 4.83 & 0.89 & $-$3.02 \\
\midrule
\multirow{6}{*}{OWID-T2V} & Co-GPT-OSS-20B & 0.60 & \textbf{22s} & 0.56 & 0.37 & 38.74 & 4.66 & 0.92 & $-$1.18 \\
 & Cl-Qwen3.6-35B & 0.90 & 228s & 0.89 & 0.54 & 64.74 & 4.67 & 0.87 & $-$3.29 \\
 & CL-Gemma4-31B & 0.99 & 132s & 0.95 & 0.62 & 72.29 & 4.85 & 0.88 & $-$1.64 \\
 & LIDA-GPT-5 & 0.79 & 132s & 0.77 & 0.52 & 54.37 & 4.60 & 0.92 & $-$2.17 \\
 & Co-GPT-5 & \textbf{1.00} & 78s & \textbf{0.97} & 0.68 & 73.67 & 4.74 & 0.90 & $-$2.07 \\
 & Cl-Opus-4.7 & \textbf{1.00} & 63s & \textbf{0.97} & \textbf{0.69} & \textbf{78.80} & 4.76 & 0.90 & $-$3.58 \\
\bottomrule
\end{tabular}
\caption{Performance of text-to-vis systems. "Co", "CL", and "LIDA" denote Codex, Claude Code, and LIDA, respectively. Readability and accessibility (text contrast, text spacing) are computed only over successfully generated chart images; the number of images therefore varies across systems according to their task completion rate.}
\label{tab:performance-of-coding-agents}
\end{table*}

\begin{table*}[!ht]
\centering
\footnotesize
\setlength{\tabcolsep}{3pt}
\begin{tabular}{llrrrrrrrr}
\toprule
\textbf{Subset} & \textbf{Model} & \textbf{Task} & \textbf{Pred.} & \textbf{Chart} & \textbf{Data} & \textbf{Correct-} & \textbf{Reada-} & \textbf{Text} & \textbf{Text} \\
 &  & \textbf{Compl.} & \textbf{Time} & \textbf{Type} & \textbf{Sim.} & \textbf{ness} & \textbf{bility} & \textbf{Contrast} & \textbf{Spacing} \\
\midrule
\multirow{6}{*}{\shortstack[l]{arXiv-T2V-easy\\\small (456 samples)}} & Co-GPT-OSS-20B & 0.66 & \textbf{23s} & 0.62 & 0.53 & 45.75 & 4.60 & 0.89 & $-$0.82 \\
 & Cl-Qwen3.6-35B & 0.98 & 215s & 0.93 & 0.80 & 73.84 & 4.73 & 0.88 & $-$3.31 \\
 & CL-Gemma4-31B & 0.99 & 102s & \textbf{0.95} & 0.84 & 75.34 & 4.74 & 0.86 & $-$0.95 \\
 & LIDA-GPT-5 & 0.97 & 137s & 0.93 & 0.84 & 75.02 & 4.55 & 0.88 & $-$0.97 \\
 & Co-GPT-5 & \textbf{1.00} & 66s & \textbf{0.95} & 0.85 & 76.28 & 4.72 & 0.89 & $-$1.30 \\
 & Cl-Opus-4.7 & \textbf{1.00} & 38s & 0.93 & \textbf{0.86} & \textbf{79.67} & 4.84 & 0.89 & $-$2.61 \\
\midrule
\multirow{6}{*}{\shortstack[l]{arXiv-T2V-hard\\\small (349 samples)}} & Co-GPT-OSS-20B & 0.69 & 97s & 0.61 & 0.43 & 37.55 & 4.46 & 0.90 & $-$1.13 \\
 & Cl-Qwen3.6-35B & 0.97 & 251s & 0.90 & 0.63 & 60.51 & 4.65 & 0.88 & $-$3.76 \\
 & CL-Gemma4-31B & 0.99 & 153s & \textbf{0.92} & 0.69 & 64.87 & 4.69 & 0.87 & $-$1.86 \\
 & LIDA-GPT-5 & 0.91 & 107s & 0.81 & 0.61 & 58.86 & 4.55 & 0.87 & $-$2.03 \\
 & Co-GPT-5 & \textbf{1.00} & 87s & 0.91 & 0.72 & 67.21 & 4.71 & 0.87 & $-$2.84 \\
 & Cl-Opus-4.7 & \textbf{1.00} & \textbf{76s} & 0.90 & \textbf{0.73} & \textbf{73.58} & 4.83 & 0.89 & $-$3.50 \\
\bottomrule
\end{tabular}
\caption{Performance on easy and hard splits of arXiv-T2V.}
\label{tab:performance_of_two_agents_on_arxiv_splits}
\end{table*}

\subsection{Execution Performance}
\label{sec:execution-performance}

\textbf{Basic execution is largely solved, except for the small local model.}
As shown in Table~\ref{tab:performance-of-coding-agents}, task completion and chart-type accuracy are near-ceiling for every strong system, completing $99$--$100\%$ of samples and chart-type accuracy between $0.90$ and $0.97$. 
The exception is Co-GPT-OSS-20B, a small local model, which completes only $76\%$, $67\%$, and $60\%$ of GitHub-T2V, arXiv-T2V, and OWID-T2V, respectively.
This means task completion remains discriminating for small local models, which still fail to reliably produce valid visualisations.

\textbf{Displayed data similarity remains far from solved.}
Displayed data similarity and correctness are the metrics that separate systems, and here every system falls well short of the ceiling.
The best score across all data subsets is $0.81$ (Cl-Opus-4.7 on arXiv-T2V), and all systems degrade sharply on OWID-T2V, where the top score is only $0.69$.

\textbf{Frontier models separate on correctness and efficiency.}
Although Co-GPT-5 and Cl-Opus-4.7 complete all tasks, Cl-Opus-4.7 achieves consistently higher correctness, improving by around 5 points across the three subsets, while also running 20-43\% faster.
Cl-Opus-4.7 matches or surpasses Co-GPT-5 in quality while achieving substantially lower latency.

\textbf{System design can matter as much as the underlying model.}
Although GPT-5 powers both Codex and LIDA, their performance diverges sharply on harder subsets. 
On OWID-T2V, LIDA-GPT-5 completes only $79\%$ of tasks compared with Co-GPT-5’s $100\%$, with lower displayed data similarity ($0.52$ vs.\ $0.68$) and correctness ($54.37$ vs.\ $73.67$). 
This shows that the surrounding pipeline---prompt construction, code execution, and iterative refinement---can shape outcomes as much as the model itself, exposing a dimension that LLM-only benchmarks miss.

\textbf{Open models rival frontier systems on easy charts but fail on hard ones.}
Cl-Gemma4-31B is the strongest system on GitHub-T2V (correctness $75.41$; displayed data similarity $0.70$).
It stays close on arXiv-T2V, but underperforms on OWID-T2V; Cl-Qwen3.6-35B follows the same trajectory.
These results shows our subsets vary in difficulty, with OWID-T2V the most discriminative for displayed data similarity.

\textbf{Latency remains a practical challenge.}
Even the fastest quality systems take tens of seconds per sample, and several are considerably slower: Cl-Qwen3.6-35B requires up to $231$s on arXiv-T2V, and the multi-stage LIDA-GPT-5 exceeds $110$s on every subset. 
From a user-experience perspective, this latency is a meaningful obstacle to interactive data exploration, and it is a dimension on which systems with comparable output quality still differ substantially.

\subsection{Readability \& Accessibility Evaluation}

Across all systems, readability scores exceed 4.5 out of 5, indicating consistently readable charts, with Opus-4.7 scoring highest.
Text contrast is likewise high for every system, at roughly 0.86--0.92 on a $[0, 1]$ scale.
Text spacing, however, is negative for all systems, from $-3.58$ to $-0.33$ on a $(-\infty, \infty)$ scale, where scores above zero denote adequate spacing~\citep{miranda-pena-etal-2024-llms}. 
This shows that generated charts tend to pack text more tightly than recommended.
Since this metric sums the per-word-pair spacing over the whole image, systems that render more text (e.g., the Claude Code pairings) accumulate more negative scores, and values worsen on the denser arXiv-T2V and OWID-T2V subsets.
% Note that readability and accessibility are computed only over successfully generated images, so the effective sample size varies across systems in proportion to their task completion rate.

%TODO
% Based on our manual inspection, this difference primarily arises from simpler chart layouts and fewer visual elements, which reduce text crowding and improve contrast, despite occasional inaccuracies in data content.

\subsection{Performance on Easy and Hard Splits of arXiv-T2V}
We further evaluate all systems on the easy and hard difficulty splits of arXiv-T2V.
The easy split contains samples where the input data is already structured and ready for direct visualisation, while the hard split requires additional preprocessing steps such as data transformation, simulation, or reformatting prior to plotting.
The assignment of samples to the easy and hard splits was decided by expert annotators.
As shown in Table~\ref{tab:performance_of_two_agents_on_arxiv_splits}, all systems exhibit degraded performance on the hard split, reflected by consistently longer prediction time and lower scores in chart type accuracy, displayed data similarity, chart correctness, readability, and text spacing.

\subsection{Other Results}
\textbf{Human Request vs. Machine-Generated Request}\quad 
% In addition to the original human-made natural language requests, 
We also constructed machine-generated request variants.
As shown in Table~\ref{tab:GPT-5_human_vs_machine} in Appendix~\ref{sec:appendix-human-machine-requests}, readability and accessibility metrics remain largely unaffected.
Notably, Co-GPT-5, the experimental model, exhibits significant improvements under machine-generated requests for GitHub-T2V and arXiv-T2V.
Because machine-generated requests tend to be more explicit, they can inflate performance on those subsets; human-authored requests offer a more realistic and challenging evaluation setting.

\textbf{Correlation Analysis}\quad 
We further conduct correlation analyses to examine the behaviour of the visualisation systems, with details in Appendix~\ref{sec:correlation-analysis}. 
Overall, we observe no significant correlations between displayed data similarity and factors such as the length of the natural language request, chart type, data source, or the length of data manipulation steps. 
Similarly, readability scores show no meaningful correlation with the number of displayed data points.

\section{Conclusion}

This paper introduces VisCodeBench, a comprehensive benchmark for evaluating text-to-vis systems from an end-to-end perspective.
VisCodeBench is comprised of three subsets, GitHub-T2V, arXiv-T2V, and OWID-T2V, which cover educational, scientific, and public-interest visualisation scenarios with varying levels of complexity. 
Each sample is designed to reflect realistic data visualisation workflows, including data loading, transformation, and chart generation with respect to human natural language requests.
% This enables a more reliable assessment of text-to-vis systems in practical settings.

We benchmarked three visualisation systems---Codex, Claude Code, and LIDA---paired with a range of open and commercial LLMs, corresponding to different deployment constraints.
We evaluated them using seven metrics: task completion rate, execution time, chart type accuracy, displayed data similarity, chart correctness, readability, and accessibility.
Our results show that strong systems now largely solve basic execution, with only the small local model struggling to reliably produce valid charts.
Content fidelity, however, remains unsolved, as no system reproduces the reference data well.
VisCodeBench separates frontier models: they differ on chart correctness score and efficiency, with ClaudeCode-Opus-4.7 gaining about 5 points in chart correctness over Codex-GPT-5 and running around 20 seconds faster, making Opus-4.7 the strongest model overall. 
Our results also show that a system's design can matter as much as its underlying model, and that latency remains a practical obstacle to interactive use.
VisCodeBench will serve as a valuable resource for advancing research on reliable and practical text-to-vis systems.

\section*{Ethics Statement}
This research was conducted under the approved ethics application "2024/HE000844" by the Research Ethics Committee at The University of Sydney. 
In conducting the study, we obtained informed consent from all participants to publish the collected data under the CC BY-SA 4.0 license. 
For external data sources such as GitHub and Our World in Data (OWID), all data used in this study clearly indicated permissive copyright and licensing terms.
Lastly, we note that users of our datasets may obtain different visualisations or results when applying their own methods, due to variations in data processing and visualisation techniques. 

\section*{Acknowledgments}
This project is partially supported by the Australian Research Council through a Discovery Early Career Researcher Award, the Commonwealth Scientific and Industrial Research Organisation (CSIRO), and an unrestricted gift from Google. 
We also acknowledge all participants of our online survey. 
Their contributions will be made available at \href{https://github.com/SAIPLL/VisCodeBench}{https://github.com/SAIPLL/VisCodeBench}.
We extend our gratitude to the anonymous reviewers for their constructive feedback and valuable advice on our submissions.

\bibliography{colm2026_conference}
\bibliographystyle{colm2026_conference}

\appendix

\section{Limitations}
% Our study has several limitations. 
% First, the dataset lacks a dedicated training split. 
% However, since our goal is to assess the zero-shot performance of LLMs, we believe this setup remains appropriate for the intended evaluation. 
\textbf{Limited evaluated models}:
Our benchmarking is limited to seven text-to-vis systems, with two proprietary LLMs: GPT-5 and Opus-4.7.
This is primarily due to practical cost constraints.
A single run on three data splits costs 175 USD for Co-GPT-5 and 780 USD for Cl-Opus-4.7.

\textbf{Static, single-turn evaluation}:
VisCodeBench evaluates a single pass—data and request in, one chart out—and does not capture the iterative, multi-agent workflows used by many modern text-to-vis systems.
We believe this setup remains meaningful and complementary: a strong system should still produce accurate results quickly in a single turn, yet even Co-GPT-5 and CL-Opus-4.7  still struggle with this setting. 

\paragraph{Reliance on LLM-assisted requests}:
For arXiv-T2V and OWID-T2V, initial requests are drafted by GPT-5 and revised by experts, which may weaken the claim that they reflect natural human phrasing.
We also release arXiv-T2V-Author, which includes original author-written requests as supplementary data.

\section{Online Survey for arXiv-T2V}
\label{sec:online-survey-arXiv}

\subsection{Target Demographic}
Our target participants are authors of papers from academic repositories (e.g., arXiv and the ACL Anthology) that are licensed under CC BY-SA 4.0, as well as researchers who can be reached through social media platforms and academic mailing lists.
To identify the most relevant contributors, we applied several filters. 
These include selecting papers that contain visual plots, having email addresses listed in the manuscript, and maintaining a balance between easily made charts and more challenging ones.
Using these selections, we identified approximately 35k target participants from 120k papers published on arXiv in 2024.

\subsection{Identifying Contactable Papers}
\label{sec:finding-contactable-paper}
% The process for finding contactable papers
% 1. Download paper's tex source from arXiv
% 2. Extracting author emails
% 3. Extracting figures in the source and classifying it into plot/not-plot
% 4. (2) + (3) --> find the contactable papers 
We employed a four-step process to identify contactable papers, including email information and plot images.
First, we downloaded the TeX repository and metadata of published computer science papers from the arXiv system in 2024~\footnote{\href{https://arxiv.org/list/cs/recent}{https://arxiv.org/list/cs/recent}}.
Second, we used several heuristics and GPT-4o-mini~\footnote{\href{https://platform.openai.com/docs/models/gpt-4o-mini}{https://platform.openai.com/docs/models/gpt-4o-mini}} to extract the first author's email address from the main TeX content, as we believe that the first authors are the primary contributors. 
Third, for papers containing the contact address, we followed the \texttt{\{figure\}} sections in the main TeX to locate attached images in the repository and classified them as plots or non-plots using GPT-4o-mini. 
Finally, papers with both an extracted email address and at least a plot image would be considered contactable papers.

Table~\ref{tab:contactable-papers-dataset-1} describes the performance of GPT-4o-mini in the email extraction task on 232 TeX repositories, and the plot classification task on 1,819 figure images, with ground truths made by our experts.
Overall, the LLMs' performance in both tasks is sufficient for our purpose of identifying contactable papers. 
Additionally, we instructed the LLM to extract only certain cases and leave uncertain ones blank for the email extraction task, resulting in 100\% precision to avoid contacting the wrong individuals.
% The prompts for both tasks can be found in our GitHub at \href{https://github.com/xxx/xxx}{https://github.com/xxx/xxx}.

Table~\ref{tab:contactable-papers-dataset-2} reports the number of papers remaining after each filtering step. 
Starting from 300 arXiv papers, 232 have an associated TeX repository; from these, we extract the first author's email for 177, and 134 of those additionally contain at least one plot image. 
A paper is considered contactable only if it has both an extractable email and at least one plot, so roughly 134 of every 300 papers (44.67\%) are contactable for data submission.

\begin{table}[t]
\centering
\begin{tabular}{cccc}
\toprule
Task & Precision & Recall & F1-score  \\ 
\midrule
Email extraction & 1.0000 & 0.9676 & 0.9835  \\ 
Plot classification & 0.9846 & 0.9846 & 0.9846  \\ 
\bottomrule
\end{tabular}
\caption{Performance of GPT-4o-mini in the email extraction task (232 repositories) and plot classification task (1,819 images).}  
\label{tab:contactable-papers-dataset-1}
\end{table}

\begin{table}[t]
\centering
\begin{tabular}{lc}
\toprule
\# Papers for experiment            & 300\\ 
\# Papers with TeX repository       & 232\\ 
\# Papers with first author's email & 177 \\ 
\begin{tabular}[c]{@{}l@{}}\# Papers with\\ \enspace  + first author's email\\ \enspace  + at least a plot image\end{tabular} & 134 \\
\bottomrule
\end{tabular}
\caption{The number of papers through each filtering step.}
\label{tab:contactable-papers-dataset-2}
\end{table}

% \subsection{Identify challenging charts}

% \subsection{Remove duplication charts}

\subsection{Platform and Distribution}
We used a dedicated email address to communicate with participants throughout the study. 
Microsoft Outlook was used to manually respond to inquiries, while an SMTP server configured with AWS SES enabled the automated distribution of mass invitation emails.
Each invitation included a brief overview of the project, information about the benefits of participation, and a link to a REDCap survey that contained a consent form and a data submission form.
% The template for the email, consent form, and data submission form can be found in our repository.

We conducted the survey in two rounds. In the first round, we sent invitation emails to approximately 5,678 people who were the first authors of 2024 arXiv papers in the Computation and Language (cs.CL) category. 
In the second round, we contacted around 25,842 people who were the first authors of papers in other computer science categories, such as cs.CV (Computer Vision and Pattern Recognition), cs.LG (Machine Learning), and cs.AI (Artificial Intelligence). 
To further increase participation, we also promoted the project via social media and academic mailing lists within and beyond our university. 

Each round allowed participants around 34 days to respond, with the option to withdraw their submitted data at any time before the dataset’s release. 
To encourage participation, two reminder emails were sent: the first two weeks after the initial invitation, and the second three to five days before the survey deadline. 
To maximise response rates, reminders were scheduled for Tuesday, Wednesday, or Thursday mornings~\citep{doi:10.1177/1525822X17715865}.
As a result, these reminders significantly boosted the submission rates in two rounds of the survey, as shown in Figure~\ref{fig:submission_by_datanumber}.

Regarding rewards for each round of the survey, the first 50 participants with successful submissions, providing correct and relevant information, received a A\$20 Amazon gift card. 
Additionally, each successful submission entered the participant into a lucky draw, where ten winners each received a A\$100 Amazon gift card.

Study data were collected and managed using REDCap electronic data capture tools hosted at The University of Sydney~\citep{HARRIS2009377, HARRIS2019103208}. REDCap (Research Electronic Data Capture) is a secure, web-based software platform designed to support data capture for research studies, providing 1) an intuitive interface for validated data capture; 2) audit trails for tracking data manipulation and export procedures; 3) automated export procedures for seamless data downloads to common statistical packages; and 4) procedures for data integration and interoperability with external sources.

\subsection{Responses}
\begin{figure}[t!]
\centering
\includegraphics[width=0.7\textwidth]{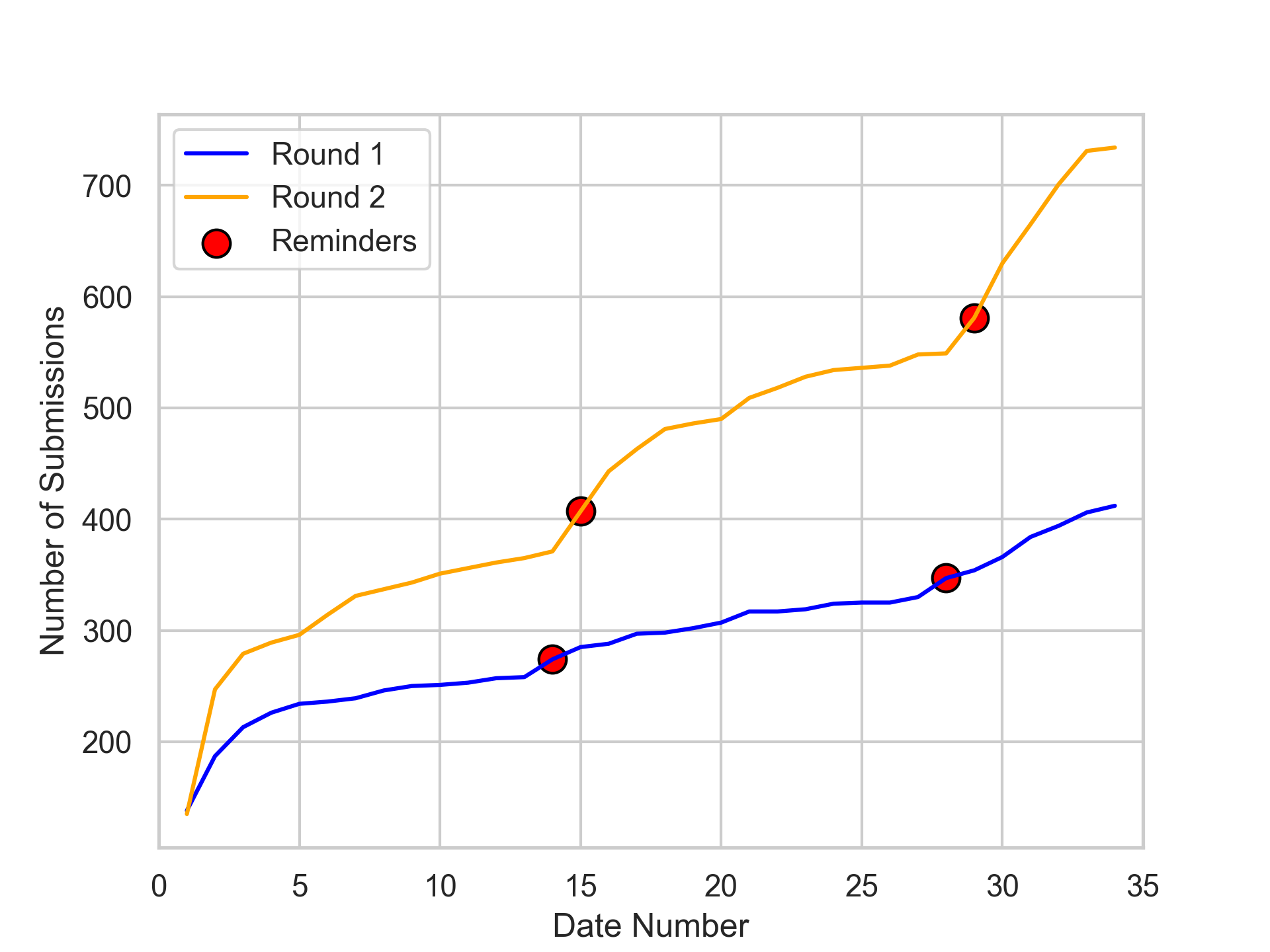}
\caption{\# Submissions by date} 
\label{fig:submission_by_datanumber}
\end{figure}

Figure~\ref{fig:submission_by_datanumber} shows the number of submissions collected via the email channel, where invitation emails were sent directly to participants.
We received submissions from 304 participants in the first round and 544 in the second, corresponding to response rates of approximately 5.3\% and 2.1\%, respectively\footnote{We count only participants who completed data submission; responses without submitted data are excluded.}.
In contrast, the social media channel, where invitations were posted on social media platforms and academic mailing lists, produced only four submissions.
In total, we collected 1,151 valid submissions, each of which includes underlying data files, a natural language request for visualisation, code files, and a resulting visualisation image.
Note that a single participant may contribute multiple submissions, with each one corresponding to a different visualisation in their papers.

\subsection{Recruitment of Expert Annotators}
We recruited two expert annotators from an NLP course at our university after the course had concluded. The recruitment process included a brief demographic questionnaire and a qualification task, which attracted 32 applicants. 
From these, we selected two candidates with over four years of coding experience and strong, well-motivated interest in the project. 
One annotator was a Master’s student with experience in both academic and industry settings, while the other was a PhD student with several years of teaching experience. 
Annotators were paid at a rate of approximately A\$49.15 per hour. 
Prior to annotation, both experts carefully studied the annotation guidelines, spent one week familiarising themselves with the task, and actively asked clarification questions before proceeding with annotating the arXiv dataset.

\section{Comparison with real-world code}
\label{sec:appendix-analysis2-realworld-comparison}
For this comparison, we focused exclusively on the Matplotlib library, building on the data used in our prior work~\citep{nguyen-etal-2024-text}, Matplotlib-nb. 
Matplotlib also dominates our datasets, appearing in 95\% of the samples. 
This focus is further supported by the fact that both Pandas and Seaborn build on Matplotlib, resulting in similar rendering behaviour across these libraries.

\begin{figure}[t!]
\centering
\includegraphics[width=0.7\textwidth]{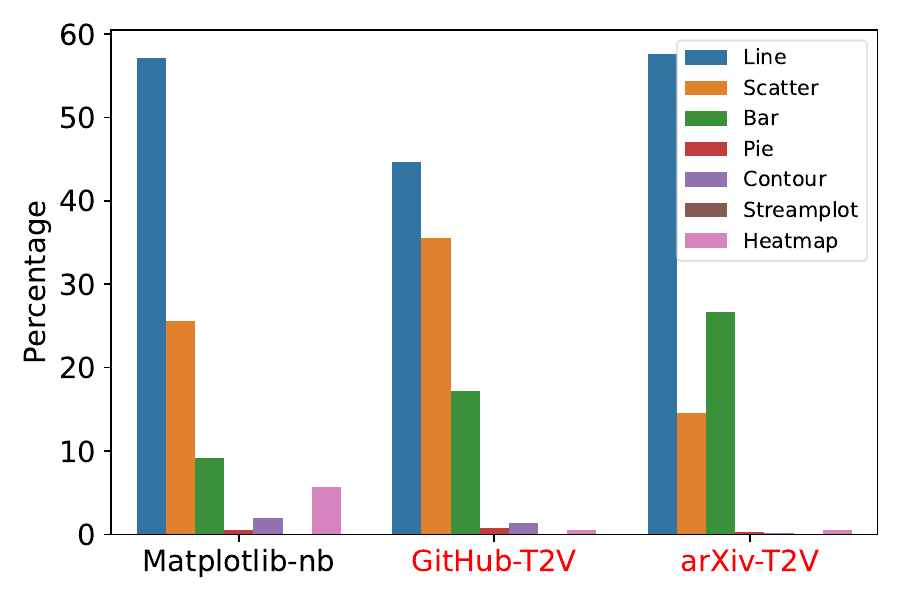}
\caption{Plot type distribution over three datasets}
\label{fig:analysis2-plottypes-comparison}
\end{figure}

\paragraph{Comparison of Chart type}
As shown in Figure~\ref {fig:analysis2-plottypes-comparison}, our dataset subsets exhibit the same trend as the real-world code.
Line plots are the most frequently used chart type across all sources, accounting for approximately 40\% to 50\% of the compared data. 
Scatter plots are the second most common type, accounting for approximately 20\% to 30\%.
Heatmaps, although present in real-world notebooks (~6\%), are rarely used in either benchmark subset.
Other chart types, such as pie, contour, and stream plots, appear infrequently in all compared data.

\begin{figure}[t!]
\centering
\includegraphics[width=0.7\textwidth]{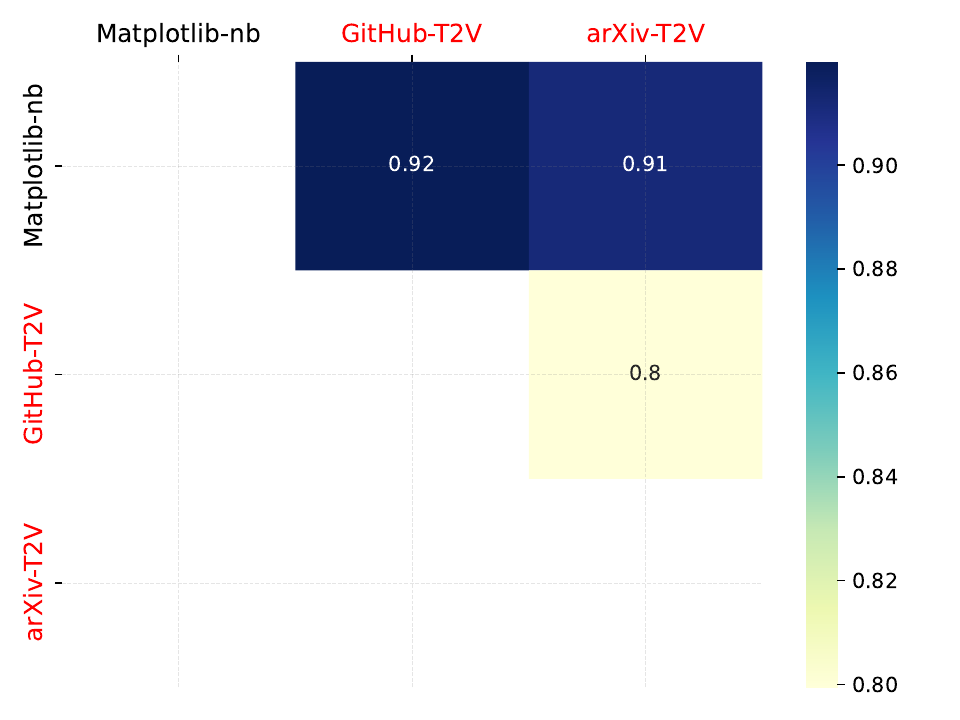}
\caption{Spearman’s rank correlation coefficient in
terms of frequent attributes}
\label{fig:analysis2-attributes-correlation-heatmap}
\end{figure}

\paragraph{Comparison of Attributes}
Figure~\ref{fig:analysis2-attributes-full} presents the normalised frequency of 62 visualisation attributes, with darker colours indicating higher usage.
Overall, our dataset subsets show strong overlap in common aesthetic and layout attributes, including titles, axis limits, tick labels, colours, opacity, line styles, annotations, legends, and the use of multiple subplots.
Furthermore, as shown in Figure~\ref{fig:analysis2-attributes-correlation-heatmap}, Spearman’s rank correlation coefficients indicate a high degree of similarity between our datasets and real-world visualisation code, with correlation scores exceeding 0.91.

\paragraph{Comparison of Complexity}
Table~\ref{tab:analysis2-complexity} reports the average number of distinct visualisation functions and parameters per code file. 
GitHub-T2V exhibits a similar level of complexity to real-world code, with averages of 6.39 functions and 9.18 parameters. 
Notably, arXiv-T2V demonstrates the highest complexity, averaging 9.20 functions and 17.32 parameters per program.

\begin{table}[t!]
\centering
\small
\begin{tabular}{ l c c}
\toprule
            & No. Funcs &  No. Params \\ \midrule
Matplotlib-nb  & 6.19          & 10.54          \\ 
\textcolor{red}{GitHub-T2V}    & 6.39          & 9.18           \\ 
\textcolor{red}{arXiv-T2V}    & \textbf{9.20}          & \textbf{17.32}           \\ 
\bottomrule
\end{tabular}
\caption{Average number of functions and parameters.}
\label{tab:analysis2-complexity}
\end{table}

\section{Extraction of Displayed Data Points}
\label{sec:appendix-data-extraction-metric}
\begin{figure}[!t]
\centering
\includegraphics[width=0.7\textwidth]{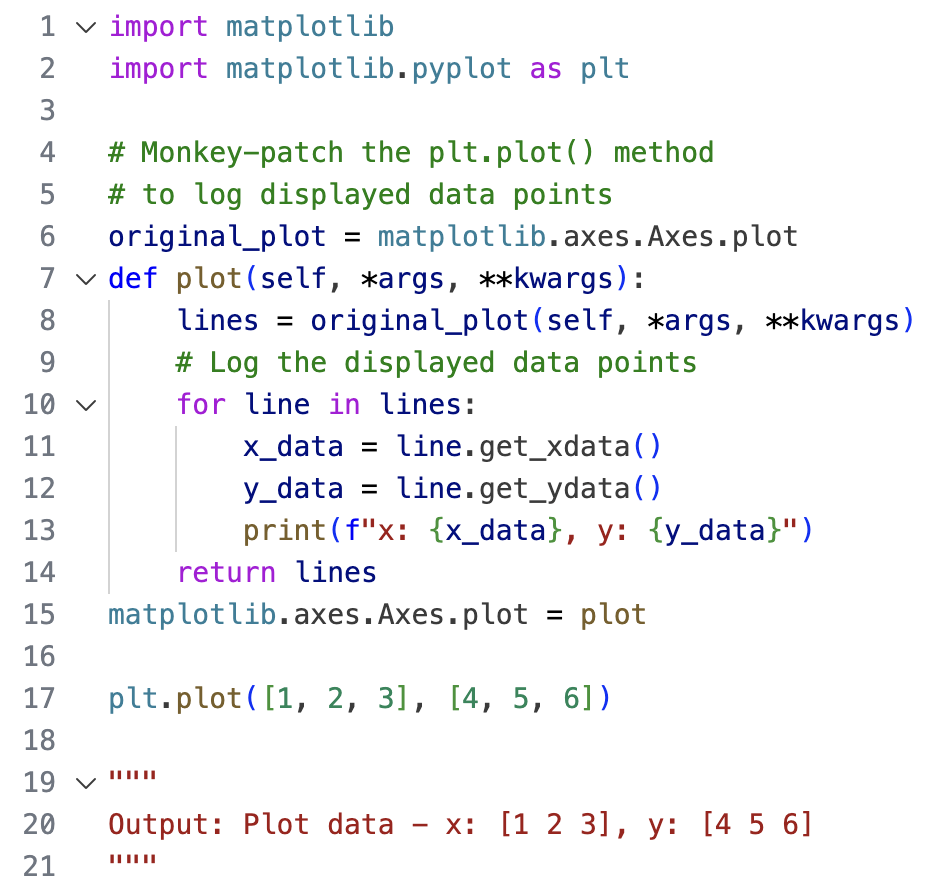}
\caption{Simple example of the monkey-patching technique}
\label{fig:monkey-patch-example}
\end{figure}

We follow a pipeline to extract displayed data points.
First, we apply a monkey-patching technique to intercept plotting calls: each time a plotting function is invoked, our logging function is automatically triggered to extract and save the displayed data points (i.e., x and y coordinates) from the axes into separate files.
Example of monkey-patch technique on \texttt{plt.plot()} function is shown in Figure~\ref{fig:monkey-patch-example}.
Table~\ref{tab:explanation-chart-type-xy-coordinates} explains the types of extracted coordinates for each chart type.

Next, we normalise the extracted data points to the range [0, 1], based on the minimum and maximum values along the horizontal and vertical axes of the subplot.
Finally, we group the normalised data points by chart type to form data series, which are then used as input for the evaluation script adopted from the CHART-Infographics competition.
This grouping is applied to all chart types except line charts, as each line requires the interpolation of data points in the evaluation script.

\begin{table}[t]
\centering
\small
\begin{tabular}{ll}
\toprule
\textbf{Chart Type} & \textbf{Extracted Data Description} \\ \midrule
Pie Plot & \begin{tabular}[c]{@{}l@{}}Values representing\\ pie segments\end{tabular} \\ \midrule
\begin{tabular}[c]{@{}l@{}}Line, Scatter, Stem, \\ Step, Event, Area Plots\end{tabular} & \begin{tabular}[c]{@{}l@{}}x, y coordinates\\  of plotted data points\end{tabular} \\ \midrule
Bar Plot, Histogram & \begin{tabular}[c]{@{}l@{}}x, y coordinates of \\ the center-top points of bars\end{tabular} \\ \midrule
Box Plot, Violin Plot & \begin{tabular}[c]{@{}l@{}}x, y coordinates\\  of min, max, mean, and \\ the first and third quartiles\end{tabular} \\ \midrule
Error Bar & \begin{tabular}[c]{@{}l@{}}x, y coordinates\\  of the top, middle, and\\  bottom of each error bar\end{tabular} \\ \midrule
\begin{tabular}[c]{@{}l@{}}2D Histogram, \\ Hexbin, Heatmap\end{tabular} & \begin{tabular}[c]{@{}l@{}}x, y coordinates of cells \\ with corresponding values (z)\end{tabular} \\ \midrule
Contour Plot & \begin{tabular}[c]{@{}l@{}}x, y, z coordinates \\ along contour lines\end{tabular} \\ \midrule
Geographic Map & \begin{tabular}[c]{@{}l@{}}x, y, z coordinates. \\ x, y represent country borders \\  while z represents the color magnitude\end{tabular} \\
\bottomrule
\end{tabular}
\caption{Explanation of extracting x, y coordinates for each chart type}
\label{tab:explanation-chart-type-xy-coordinates}
\end{table}

\section{Human Study on Displayed Data Similarity}
\label{sec:appendix-human-study}

To validate that the Displayed Data Similarity metric aligns with human perception, we conducted a human study comparing the metric against expert judgments.
We sampled 200 (predicted, reference) chart pairs and asked two expert annotators to independently rate, on a $0$--$9$ scale, how closely the data displayed in the predicted chart matched the reference.
Annotators viewed only the rendered chart images through a custom interface, mirroring how a person would visually assess similarity, while the automatic metric operates on the underlying extracted data points.
We then measured the correlation between each annotator's ratings and the automatic scores.

\begin{table}[t]
\centering
\begin{tabular}{lcc}
\toprule
 & Annotator 1 & Annotator 2 \\
\midrule
Pearson $r$    & $0.558$  & $0.415$ \\
Spearman $\rho$ & $0.557$  & $0.392$ \\
Kendall $\tau$  & $0.452$  & $0.323$ \\
Residual mean   & $-0.045$ & $-0.184$ \\
\bottomrule
\end{tabular}
\caption{Correlation between expert ratings and the automatic Displayed Data Similarity score ($n=200$). Residual mean is the average of (human $-$ automatic) scores after rescaling to a common range.}
\label{tab:human-study-ddata}
\end{table}

Table~\ref{tab:human-study-ddata} reports the results.
Both annotators show a positive and statistically significant correlation with the automatic metric ($n=200$, $p<0.001$), confirming that the metric captures the human notion of displayed-data similarity.
The correlation is moderate rather than strong, which we attribute to a difference in what each side can perceive.
The metric evaluates the full set of extracted data points and solves an optimal point-, series-, and subplot-level assignment, so it registers fine-grained discrepancies such as individual points in dense scatter plots, small deviations along long lines, and errors in secondary series or subplots.
Human annotators, judging from rendered images, perceive only coarse and salient features (overall trend, shape, and the dominant series) and cannot resolve these point-level differences.
This asymmetry also explains the negative residual mean: annotators systematically rate pairs higher than the metric, over-crediting charts whose small or hidden errors they cannot see.

\section{Setting up Visualisation Systems}
\label{sec:appendix-setting-up-coding-agents}

We benchmark three visualisation systems: two general-purpose coding agents, Codex CLI and Claude Code, and LIDA, a dedicated text-to-vis pipeline.

\paragraph{Coding agents.}
For each text-to-vis sample, we organised all required resources, including the underlying data and any dependent files, into a dedicated folder.
We then created a code template within this folder that contains the context code and a pair of special markers (\texttt{\#\#START VISUALISATION CODE} and \texttt{\#\#END VISUALISATION CODE}) indicating where the agent should insert the generated Python code.
This design is inspired by AlphaEvolve, in which a coding agent reasons over the provided context and writes code within predefined regions~\citep{novikov2025alphaevolve}.
After preparing the folder and template, we invoked the agent to complete the visualisation task with the natural-language visualisation request embedded in the prompt.
Codex CLI and Claude Code share this setup, differing only in the agent binary and the API each uses (OpenAI for Codex, the Anthropic Messages API for Claude Code).
The structure of the per-sample codebase, the Python template, and the prompt used for the coding agent are illustrated in Figure~\ref{fig:text-to-vis-sys} (left), while an example of running the agent is described on the right of the figure.
Further details of each agent's configuration and prompts can be found in our repository.

\paragraph{LIDA.}
We adapt LIDA~\citep{dibia-2023-lida} to our benchmark. Our implementation differs from the original in the kind of input it consumes: the original LIDA takes a single tabular file and lets the system decide what to visualise, whereas ours takes the same per-sample folder used for the coding agents---a scaffolded Python script with the two markers, a natural-language request, and any accompanying data files---and completes the missing visualisation code.
The three-stage design of the original pipeline is preserved: (1) \emph{data summarisation}, where the model probes the sample folder by writing and executing a short exploration script whose output serves as the summary, rather than profiling a single dataframe; (2) \emph{goal}, which is read verbatim from the user's request rather than invented by the model; and (3) \emph{visualisation generation}, where the model completes the code between the two markers and the full script is executed to produce the chart images.
The underlying idea is therefore identical to the original---visualisations are treated as executable code that must run, be inspectable, and be repairable---while the surrounding data handling is made flexible enough to accommodate the free-form, multi-file samples in our benchmark.

\paragraph{Model deployment.}
We accessed the commercial models (GPT-5, Opus-4.7) through their providers' APIs; for example, GPT-5 was served via the OpenAI portal \href{https://api.openai.com/v1}{https://api.openai.com/v1}.
The open models (GPT-OSS-20B, Gemma-4-31B, Qwen-3.6-35B) were deployed locally on a machine whose GPU has sufficient memory to support the model's maximum context length. For instance, GPT-OSS-20B was served with Ollama on a RunPod instance with an RTX 4090 GPU, supporting a maximum context length of 120k tokens.

\section{Human requests vs machine requests}
\label{sec:appendix-human-machine-requests}
In addition to GitHub-T2V, arXiv-T2V, and OWID-T2V, which include human-generated natural language requests, we created corresponding versions featuring machine-generated requests produced by GPT-5. 
The word limits for these versions were adjusted to align with their respective real-world contexts: 50 words for GitHub-T2V and OWID-T2V, and 120 words for arXiv-T2V.
These variants enable controlled comparisons of coding-agent performance under human versus machine-generated requests within the same data source.
Looking at the performance of the coding agents on readability and accessibility metrics, we observe no substantial differences between human-authored and machine-generated request settings.

In contrast, the content metrics exhibit notable gaps, as summarised in Table~\ref{tab:GPT-5_human_vs_machine}.
Co-GPT-5 improves under machine-generated requests on GitHub-T2V and arXiv-T2V, most markedly on GitHub-T2V, with relative gains of about 6\% in chart type accuracy, 43\% in displayed data similarity, and 23\% in correctness.
OWID-T2V, by contrast, shows no improvement and a slight drop in displayed data similarity, likely because its analytically complex charts require intent that a short machine-generated request cannot capture as well as a carefully written human request.
These results suggest that more explicit machine-generated requests can inflate performance on educational and scientific charts, whereas human-authored requests provide a more realistic and challenging evaluation.

\begin{table*}[t]
\centering
\footnotesize
\setlength{\tabcolsep}{4pt}
\begin{tabular}{@{}l cccccccc@{}}
\toprule
\textbf{Subset}
  & \makecell{\textbf{Task}\\\textbf{Compl.}}
  & \makecell{\textbf{Pred.}\\\textbf{Time}}
  & \makecell{\textbf{Chart}\\\textbf{Type}}
  & \makecell{\textbf{Disp.}\\\textbf{Data}}
  & \makecell{\textbf{Correct-}\\\textbf{ness}}
  & \makecell{\textbf{Reada-}\\\textbf{bility}}
  & \makecell{\textbf{Text}\\\textbf{Contrast}}
  & \makecell{\textbf{Text}\\\textbf{Spacing}} \\
\midrule
\multicolumn{9}{l}{\textit{GitHub-T2V}} \\
\quad Human-Req   & \textbf{0.998} & 68s          & 0.897          & 0.609          & 69.92          & \textbf{4.847} & \textbf{0.882} & \textbf{$-$0.984} \\
\quad Machine-Req & 0.991          & \textbf{52s} & \textbf{0.951} & \textbf{0.870} & \textbf{86.19} & 4.843          & 0.879          & $-$1.331          \\
\midrule
\multicolumn{9}{l}{\textit{arXiv-T2V}} \\
\quad Human-Req   & \textbf{1.000} & 75s          & 0.934          & 0.796          & 72.36          & \textbf{4.717} & 0.882          & $-$1.979          \\
\quad Machine-Req & 0.998          & \textbf{74s} & \textbf{0.948} & \textbf{0.814} & \textbf{77.40} & 4.683          & 0.882          & \textbf{$-$1.661} \\
\midrule
\multicolumn{9}{l}{\textit{OWID-T2V}} \\
\quad Human-Req   & 0.998          & 78s          & 0.973          & \textbf{0.685} & \textbf{73.67} & 4.739          & \textbf{0.896} & \textbf{$-$2.072} \\
\quad Machine-Req & \textbf{0.999} & \textbf{75s} & 0.973          & 0.659          & 73.10          & \textbf{4.746} & 0.895          & $-$2.271          \\
\bottomrule
\end{tabular}
\caption{Performance of Co-GPT-5 with human vs.\ machine-generated requests.}
\label{tab:GPT-5_human_vs_machine}
\end{table*}

\section{Correlation Analysis}
This section examines how the displayed data similarity metric correlates with several factors, including the length of the natural language request, chart type, data source, and the number of data manipulation steps. 
It also examines the relationship between chart readability and the number of data points displayed.
We mainly present results for Co-GPT-5 across three dataset splits: GitHub-T2V, arXiv-T2V, and OWID-T2V.

\label{sec:correlation-analysis}

\subsection{Analysis 1 - NL Request Length vs Displayed Data Similarity}
\begin{figure}[t!]
    \centering
    \includegraphics[width=0.8\linewidth]{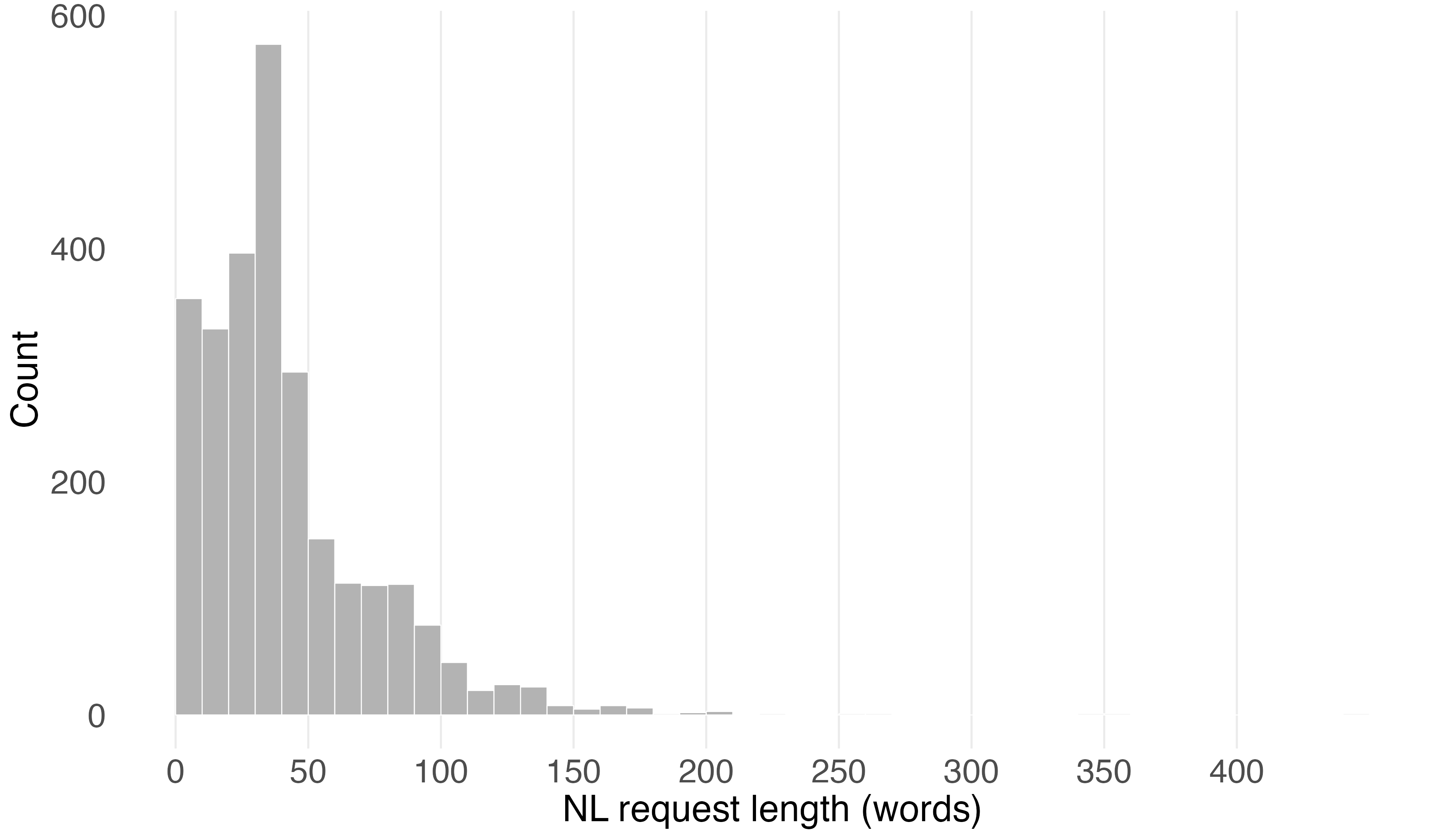}
    \caption{Histogram of human-generated natural language request length (in words) pooled across GitHub-T2V, arXiv-T2V, and OWID-T2V.}
    \label{fig:nl_request_length_pooled}
\end{figure}

\begin{figure}[t!]
    \centering
    \includegraphics[width=\linewidth]{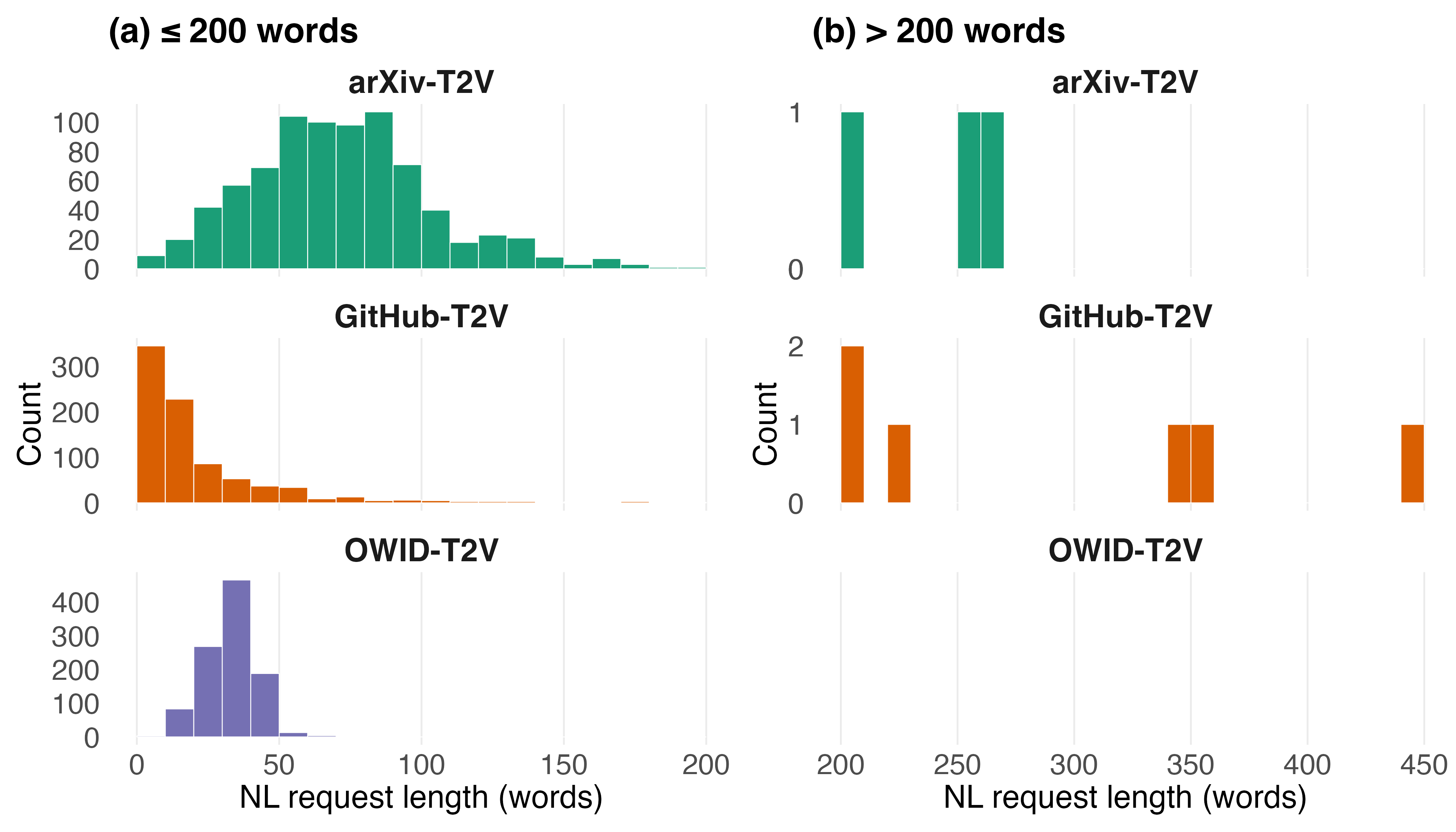}
    \caption{Histogram of human-generated natural language request length (in words) for each dataset: GitHub-T2V, arXiv-T2V, and OWID-T2V.}
    \label{fig:nl_request_length_each_dataset}
\end{figure}

Figure~\ref{fig:nl_request_length_pooled} presents a histogram of the length in words of the natural language requests pooled across the three subsets: GitHub-T2V, arXiv-T2V, and OWID-T2V. The distribution is right-skewed, with most requests around 50 words in length. 
Figure~\ref{fig:nl_request_length_each_dataset} presents histograms of the length in words of these requests broken down by dataset. Separate panels show a) the bulk of requests ($\leq 200$ words), and b) the few long requests ($>200$ words), which were present in the GitHub-T2V and arXiv-T2V.

The length of the bulk of requests is distributed differently across the datasets. 
The arXiv-T2V requests tend to be longer, with wider variance in the request length.
GitHub-T2V requests are skewed to shorter request lengths, with most requests shorter than ~60 words. 
The OWID-T2V requests are tightly clustered around 30-40 words.

\begin{figure}[t!]
    \centering
    \includegraphics[width=\linewidth]{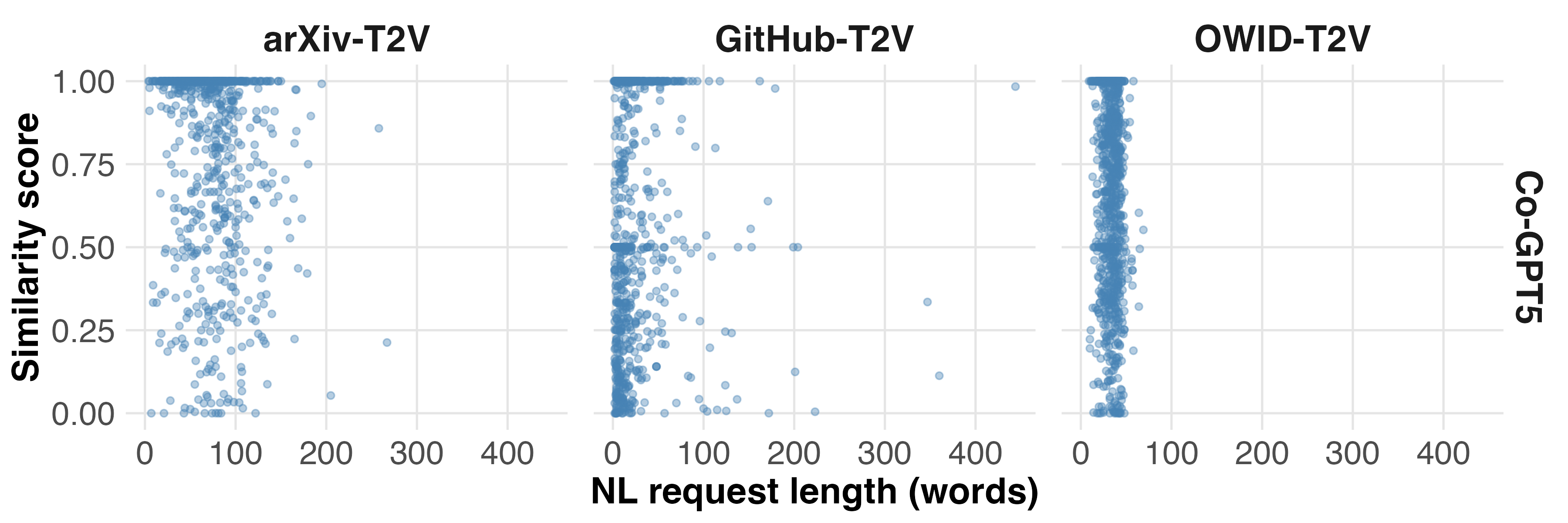}
    \caption{Scatter plots of similarity score versus NL request length for each dataset-agent combination (GitHub-T2V, arXiv-T2V, OWID-T2V; Co-GPT-5)}
    \label{fig:nl_request_length_vs_similarity_score}
\end{figure}

Natural language request length was not meaningfully associated with similarity score for any of the runs investigated. 
For the arXiv-T2V subset, we found a non-significant Pearson coefficient, indicating no linear association between request length and similarity score ($r$ = -0.19, $p$ = 3.76e-08). 
The Spearman coefficient, while significant, was close to zero ($\rho$ = -0.225, $p$ = 1.16e-10), suggestive of only a weak monotonic association. 
This pattern was replicated across the other combinations of subsets and agents.
There was thus no clear or consistent relationship between natural language request length and similarity score. 
Scatter plots for similarity score versus NL request length in words for each dataset-agent combination are presented in Figure~\ref{fig:nl_request_length_vs_similarity_score}.

\subsection{Analysis 2 - Data Source Type vs Displayed Data Similarity}
\label{ssec:appendix-correlation-analysis-2}

In the arXiv-T2V and GitHub-T2V examples, the data for plotting is accessible from different sources. These sources are broadly separated into four categories:

\begin{itemize}
    \item File: Where the plotting code reads in an external file (e.g., a .csv or .json file), which contains the data for plotting.
    \item Manual: Where the data for plotting is manually specified/embedded into the code file (e.g., a list or dictionary of values).
    \item Simulation: Where the data to plot is simulated using random numbers or some mathematical function is used to generate the data.
    \item Other: Where a plot falls into more than one of the above categories (e.g., a figure with two subplots, where one subplot falls under the `File' category and the other falls under the `Simulation' category), or none of the above categories (e.g., the data for plotting comes from a library's built-in datasets, such as Scikit-learn’s Iris dataset).
\end{itemize}

\begin{figure}[t]
    \centering
    \begin{tcolorbox}[
        colback=gray!10, 
        colframe=gray!50, 
        width=\linewidth,        % make the box full text width
        fontupper=\small,        % make the text inside smaller
        sharp corners,
        boxrule=0.5pt            % optional: thinner frame
    ]

You will be given a code snippet to generate a plot. Your task is to act as a classifier to determine what source of data is used for the underlying plot data. You can choose four options:

\vspace{1.5mm}

- "File" for when the original data to plot is stored in a file. For example, the data for plotting is predominantly stored in a file (e.g., the data is read in from a file "data  = open('file', 'r')"). It should still be considered the "File" class where a file is loaded in, and then processing is done to the file's contents. It should also be considered the "File" class where most of the data is in the file, but some extra data is manually appended.

\vspace{1.5mm}

- "Simulation" for when the data to plot is simulated or randomly generated. This class is assigned where mathematical functions are applied to generate the data (e.g., we start with torch.ones((1, 64, dim)) and then apply calculations). It could also involve random number generation.

\vspace{1.5mm}

- "Manual" for when the data to plot is provided directly in the file. For example, the data for plotting may be provided in lists (e.g. x = [1,2,3,4,5], y = [-1,-2,-3,-4,-5]) or dictionaries. Where the data is manually specified, but simulation is done using that data, it is classified as the "Simulation" class.

\vspace{1.5mm}

- "Other" when you believe the other class labels are inappropriate. In this case, you must also provide a brief explanation of why it does not fit under the "Other" classifications. An example is where there are two subplots, and one fits the "File" class and the other the "Manual" class. 

\vspace{1.5mm}

Only return the class label, and nothing else. If it is the "Other" label, write "Other" first, followed by your reason. Please think carefully before providing your answer.
    \end{tcolorbox}
    \caption{GPT-5 prompt used for the data source classification in ~\ref{ssec:appendix-correlation-analysis-2}}
    \label{fig:gpt_prompt_data_source}
\end{figure}

The OWID-T2V data was not included in this analysis as its data always comes from a file.

To classify the data examples, we prompted GPT-5 (prompt in Figure~\ref{fig:gpt_prompt_data_source}) and validated the approach on a small set of manually labeled samples. We found that it correctly classified all examples.
Table~\ref{tab:data_sources_proportion} presents the proportion of different data sources for the examples in each data subset. ``File'' is clearly the most common data source, with ``Manual'' and ``Other'' contributing a far smaller proportion.

\begin{table}[t]
\centering
\begin{tabular}{lcc}
\hline
\textbf{Data Source} & \textbf{arXiv-T2V} & \textbf{GitHub-T2V} \\
\hline
File       & 56.27\% & 75.09\% \\
Simulation & 6.58\% & 18.62\%\\
Manual     & 36.02\%  & 1.30\%  \\
Other      & 1.12\%  & 4.98\%  \\
\hline
\end{tabular}
\caption{Proportion of data source classifications for different datasets.}
\label{tab:data_sources_proportion}
\end{table}

\begin{figure}
    \centering
    \includegraphics[width=0.8\linewidth]{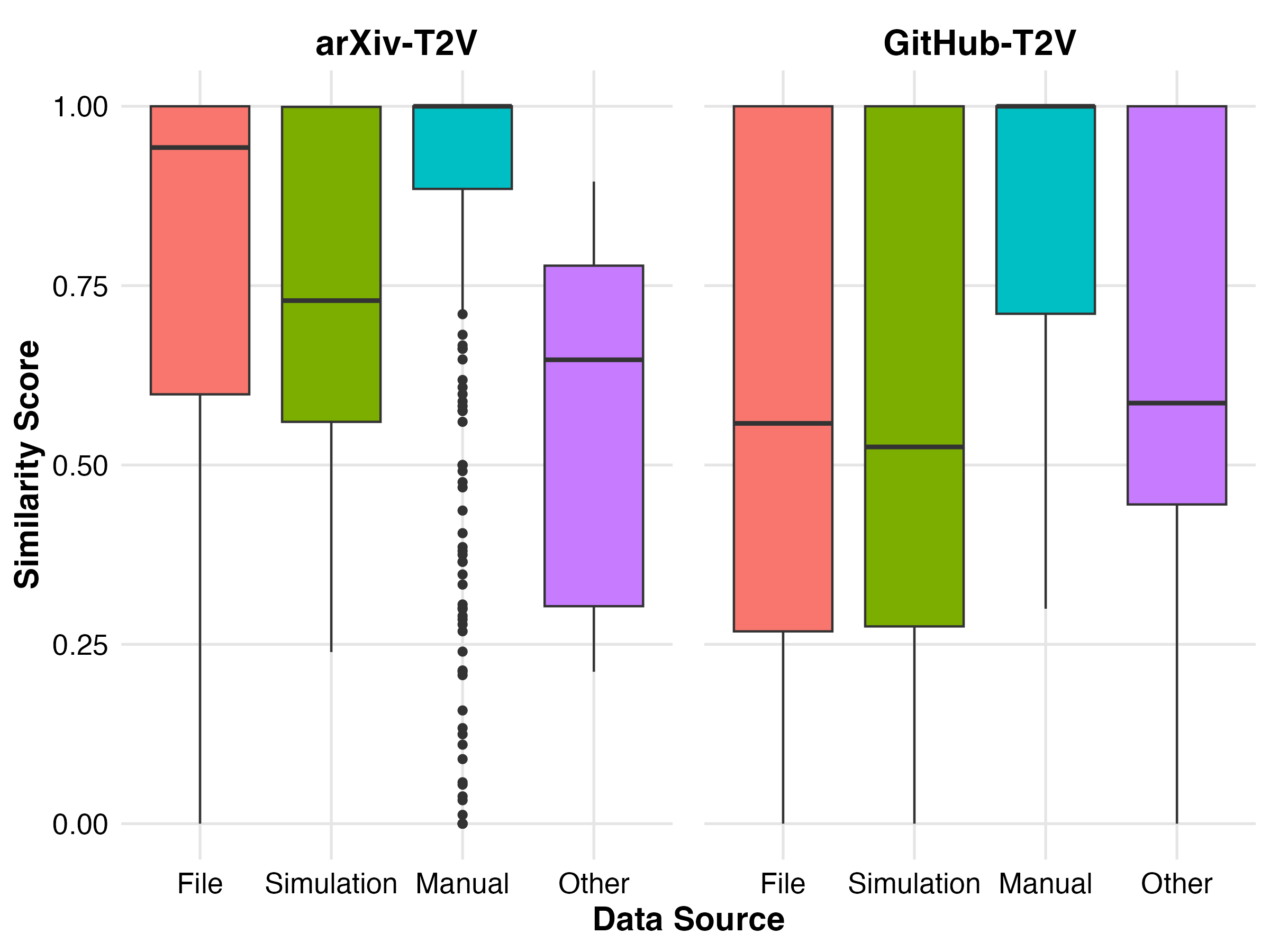}
    \caption{Similarity score vs different data sources for different data subsets separated by model configuration.}
    \label{fig:data_source_boxplot_all_data}
\end{figure}

Figure~\ref{fig:data_source_boxplot_all_data} shows the similarity scores for different data sources separated by data subsets and model type.
There is a large spread across all data source measures. 'Manual' appears to have the least variability while having the greatest performance, yet has a long left-skew. 

While 'File' has a higher median similarity score when compared to 'Simulation' and 'Other' in arXiv-T2V, this relationship does not hold with the GitHub-T2V data, where 'File', 'Simulation', and 'Other' seem to have similar medians and spreads.

\subsection{Analysis 3 - Data Manipulation vs Displayed Data Similarity}
\label{ssec:appendix-correlation-analysis-3}

In most arXiv-T2V, GitHub-T2V and OWID-T2V examples, the read-in data is manipulated before plotting. This includes, but is not limited to, data filtering, unit conversions, aggregation, merging different data frames, and applying dimension reduction (e.g., t-SNE). To quantify the number of data manipulation operations that have taken place for a given example, we prompted GPT-5 (the prompt in Figure~\ref{fig:gpt_prompt_data_manipulations}) to count the number of data manipulation operations.

\begin{figure}[t]
    \centering
    \begin{tcolorbox}[
        colback=gray!10, 
        colframe=gray!50, 
        width=\linewidth,        % make the box full text width
        fontupper=\small,        % make the text inside smaller
        sharp corners,
        boxrule=0.5pt            % optional: thinner frame
    ]

You will be given a code snippet that loads, processes, and plots data. Your task is to count the number of distinct data manipulation operations performed before plotting.

\vspace{1.5mm}

Trace the data processing steps and identify each unique data manipulation operation. Examples include:

- Filtering rows

- Type casting columns

- Unit conversions

- Renaming categories

- Data aggregation

- Merging datasets

- Appending values

- Dimensionality reduction (e.g., TSNE)

- Generating synthetic data

\vspace{1.5mm}

Do NOT count data loading (e.g., reading a CSV) as data manipulation. If the code only loads data and plots it with no intermediate processing, return 0.

\vspace{1.5mm}

If the language is not Python,  return null.

\vspace{1.5mm}

Return only your answer in the required JSON structure:

\begin{verbatim}
{
"num_steps": <integer or null>
}
\end{verbatim}

    \end{tcolorbox}
    \caption{GPT-5 prompt used to count the number of data manipulations for a given example in ~\ref{ssec:appendix-correlation-analysis-3}}
    \label{fig:gpt_prompt_data_manipulations}
\end{figure}

\begin{figure}
    \centering
    \includegraphics[width=0.9\linewidth]{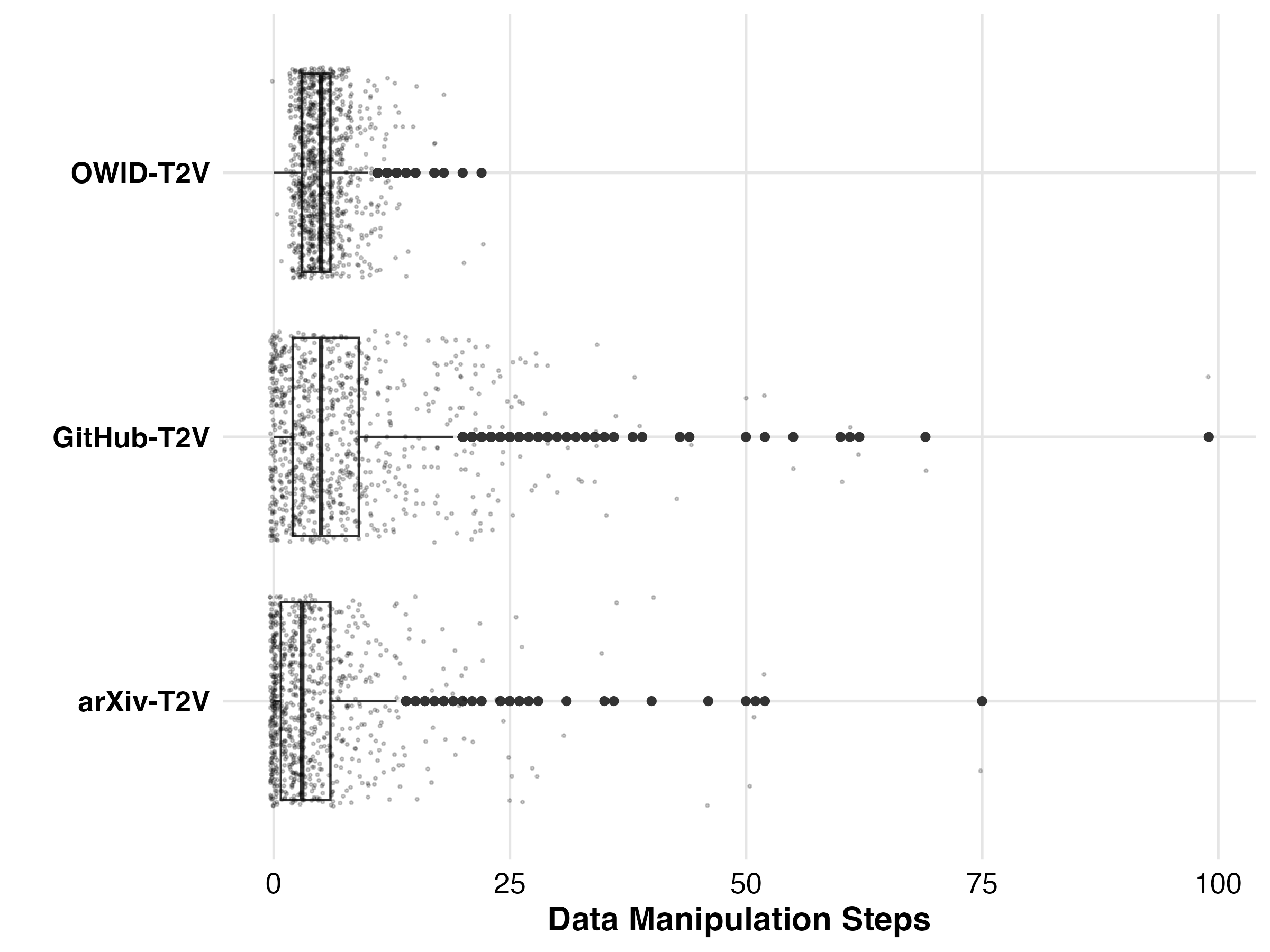}
    \caption{Number of data manipulation steps for arXiv-T2V, GitHub-T2V and OWID-T2V examples.}
    \label{fig:data_manipulation_steps}
\end{figure}

Figure~\ref{fig:data_manipulation_steps} shows the number of data manipulation steps for each example as determined by GPT-5. Combining all datasets, the average number of data manipulation steps is 5.73 ($s=$ 6.72).

\begin{figure}
    \centering
    \includegraphics[width=\linewidth]{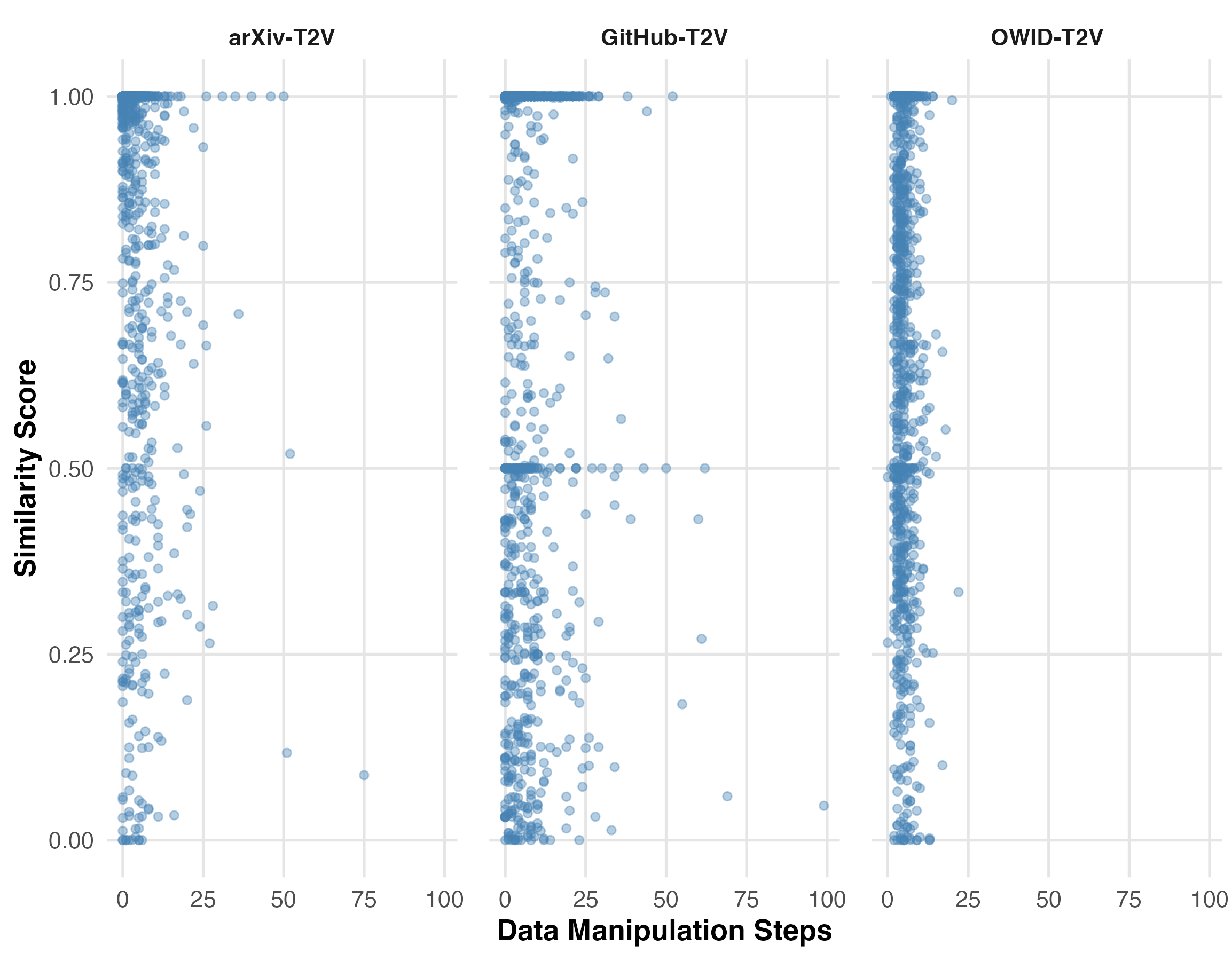}
    \caption{Data manipulation steps against similarity score for different datasets and models.}
    \label{fig:data_manipulation_linear}
\end{figure}

Figure~\ref{fig:data_manipulation_linear} shows scatter plots of data manipulation steps against the similarity score for the different data subsets and models. There appears to be no strong linear relationship between data manipulation steps and similarity score.

\subsection{Analysis 4 - Chart Type vs Displayed Data Similarity }
\begin{figure}[t!]
    \centering
    \includegraphics[width=\linewidth]{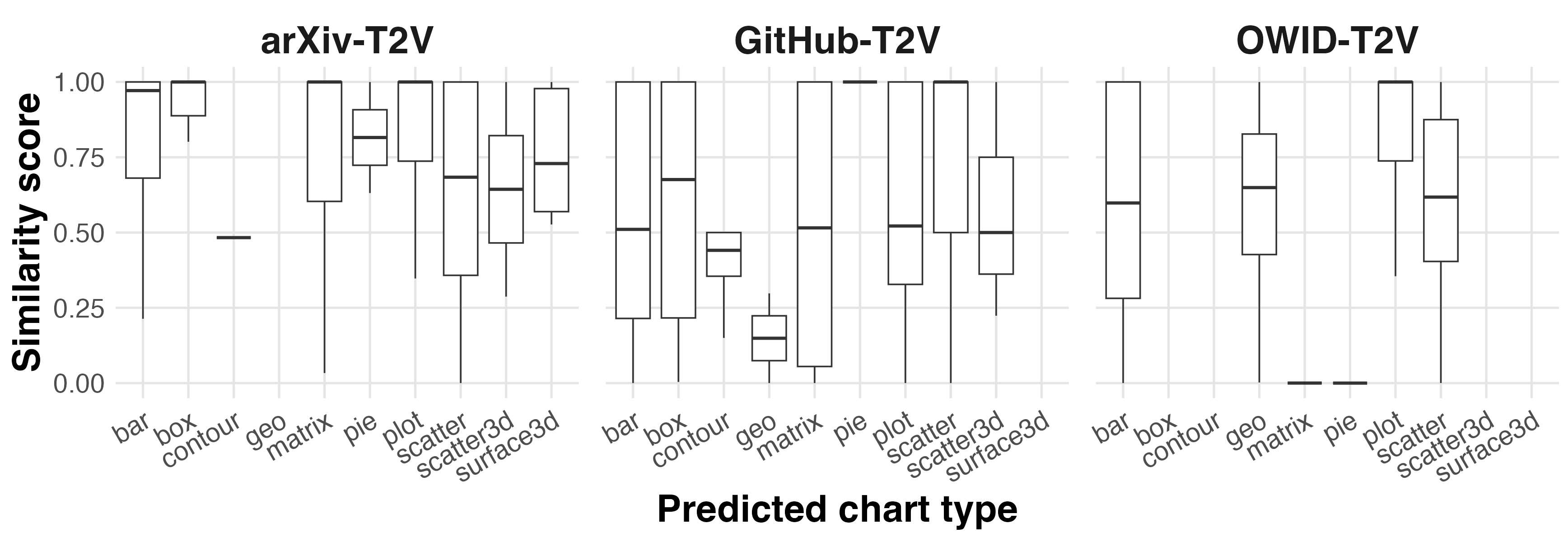}
    \caption{Similarity score across different chart types for each dataset-agent combination (GitHub-T2V, arXiv-T2V, OWID-T2V; Co-GPT-5)}
    \label{fig:sim_score_chart_type}
\end{figure}

\paragraph{Similarity vs Primary Chart type}
Across all three dataset-agent combinations tested with the Co-GPT-5 agent, a Kruskal-Wallis test found significant differences in similarity score across chart types. 
For the GitHub-T2V subset ($N = 842$), $H(6) = 35.07$, $p = 4.17 \times 10^{-6}$;
for the arXiv-T2V subset ($N = 801$), $H(5) = 51.28$, $p = 7.57 \times 10^{-10}$;
and for the OWID-T2V subset ($N = 1021$), $H(3) = 196.83$, $p = 2.04 \times 10^{-42}$. Note that, due to sample size constraints, not all chart types pictured in Figure~\ref{fig:sim_score_chart_type} were used in statistical testing.
A similarity score that differs across primary predicted chart types indicates that some figures may be more difficult to plot clearly across datasets and models. 
As shown in Figure~\ref{fig:sim_score_chart_type}, simpler charts, such as box plots and scatter plots, tend to have higher similarity scores across datasets and models.

\begin{figure}[t!]
    \centering
    \includegraphics[width=\linewidth]{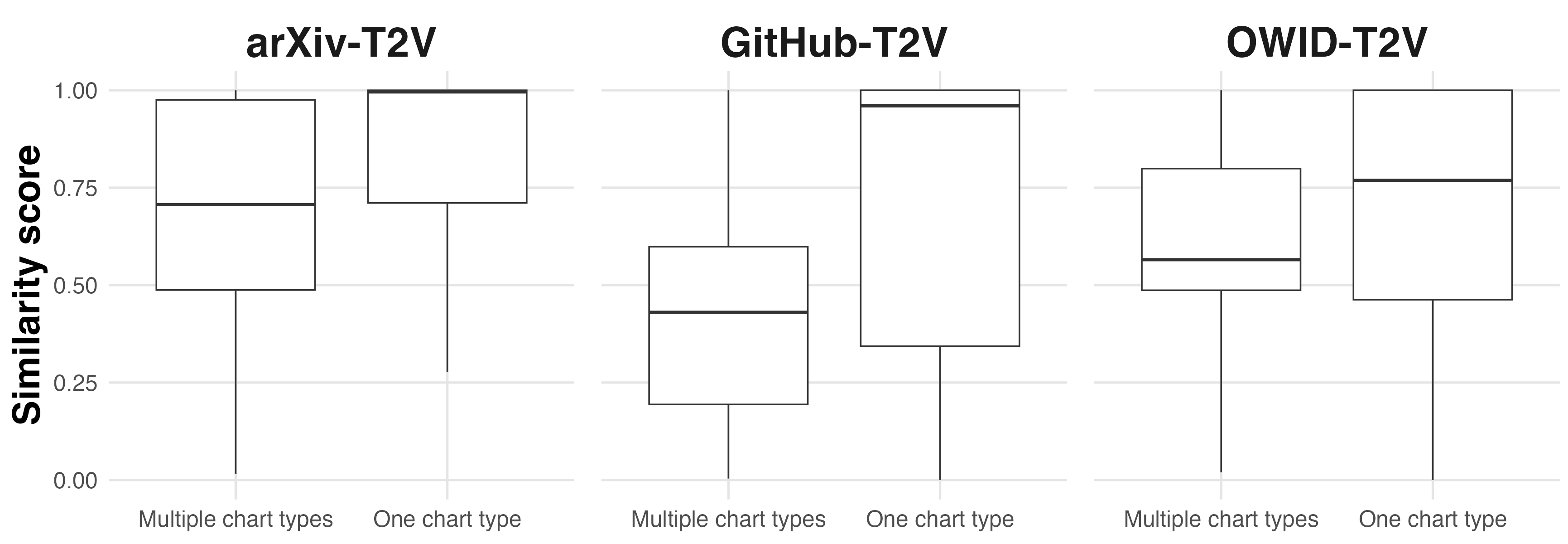}
    \caption{Similarity score for one versus multiple predicted chart types for each dataset-agent combination (GitHub-T2V, arXiv-T2V, OWID-T2V; Co-GPT-5)}
    \label{fig:sim_score_one_vs_mult_chart_type}
\end{figure}

\paragraph{Similarity for One versus Multiple Chart Types}
Figure~\ref{fig:sim_score_one_vs_mult_chart_type} compares the similarity score for visualisations with one predicted chart type to the score for visualisations with multiple predicted chart types, for the first run of each dataset-agent combination tested. 
For all experiment settings, visualisations with one chart type tended to have a higher similarity score than visualisations with multiple chart types. 
Mann-Whitney U tests found the difference between one and multiple chart types was significant for all experiment settings (for the arXiv-T2V dataset with Co-GPT-5 agent: $U=58,299,p=3.76e-10; N_\text{one}1=671, N_\text{multiple}M=130$). 
This suggests that visualisations requiring multiple chart types are more difficult to create across datasets and agents.

\begin{figure}[t!]
    \centering
    \includegraphics[width=\linewidth]{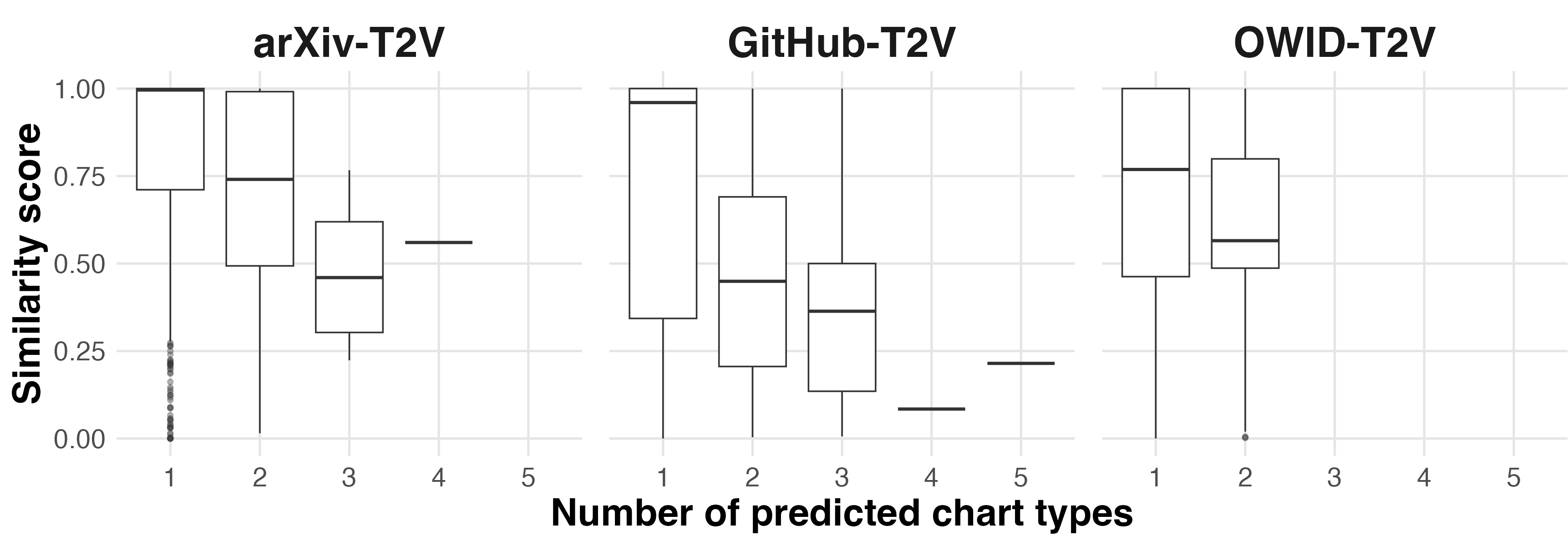}
    \caption{Similarity score versus number of chart types predicted  for each dataset-agent combination (GitHub-T2V, arXiv-T2V, OWID-T2V; Co-GPT-5)}
    \label{fig:sim_vs_num_chart_type}
\end{figure}

\paragraph{Similarity Score versus Number of Chart Types}
Figure~\ref{fig:sim_vs_num_chart_type} shows the similarity score across the number of predicted chart types, for the first run of each dataset-agent combination tested.
As the number of predicted chart types increases, the similarity score tends to decrease. 
This observation was confirmed with significant negative associations between similarity score and the number of chart types found. 
For the arXiv-T2V dataset with Co-GPT-5, both Pearson and Spearman coefficients were significant ($r=-0.193, p=3.77e-08; \rho=-0.225, p=1.16e-10$). 
This was true of all other dataset-agent combinations tested.

\subsection{Analysis 5 - Readability vs Number of Data Points}

\begin{figure}[t!]
    \centering
    \includegraphics[width=\linewidth]{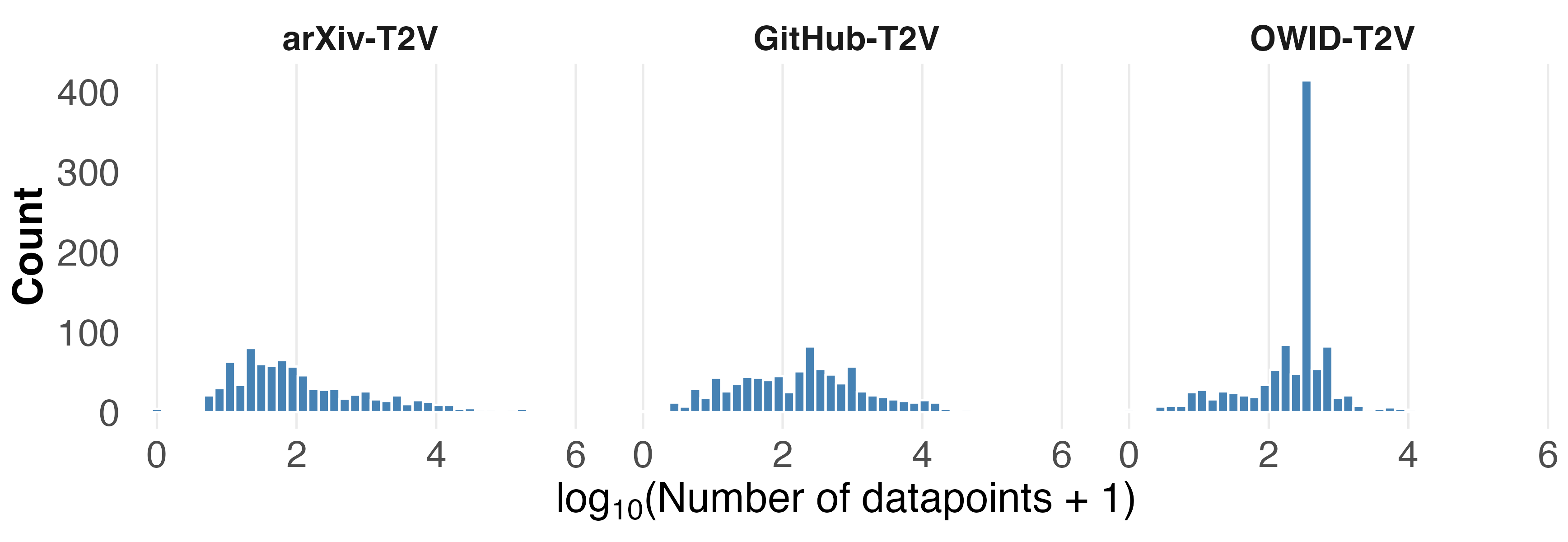}
    \caption{Logged distribution of number of data points in visualisations produced in run 1 for each dataset-agent combination (GitHub-T2V, arXiv-T2V, OWID-T2V; Co-GPT-5)}
    \label{fig:num_data_points_hist}
\end{figure}

Figure~\ref{fig:num_data_points_hist} shows the distribution of the logged number of data points in the visualisations produced in response to human natural language requests for the first run on 1) GitHub-T2V, 2) arXiv-T2V, and 3) OWID-T2V subsets using Co-GPT-5 agent. 
OWID-T2V figures tend to contain more data points than other subsets, with a sharp spike at 300 data points reaching a count of around 400, suggesting many datasets in this source have a very similar, standardized number of data points.

\begin{figure}[t!]
    \centering
    \includegraphics[width=\linewidth]{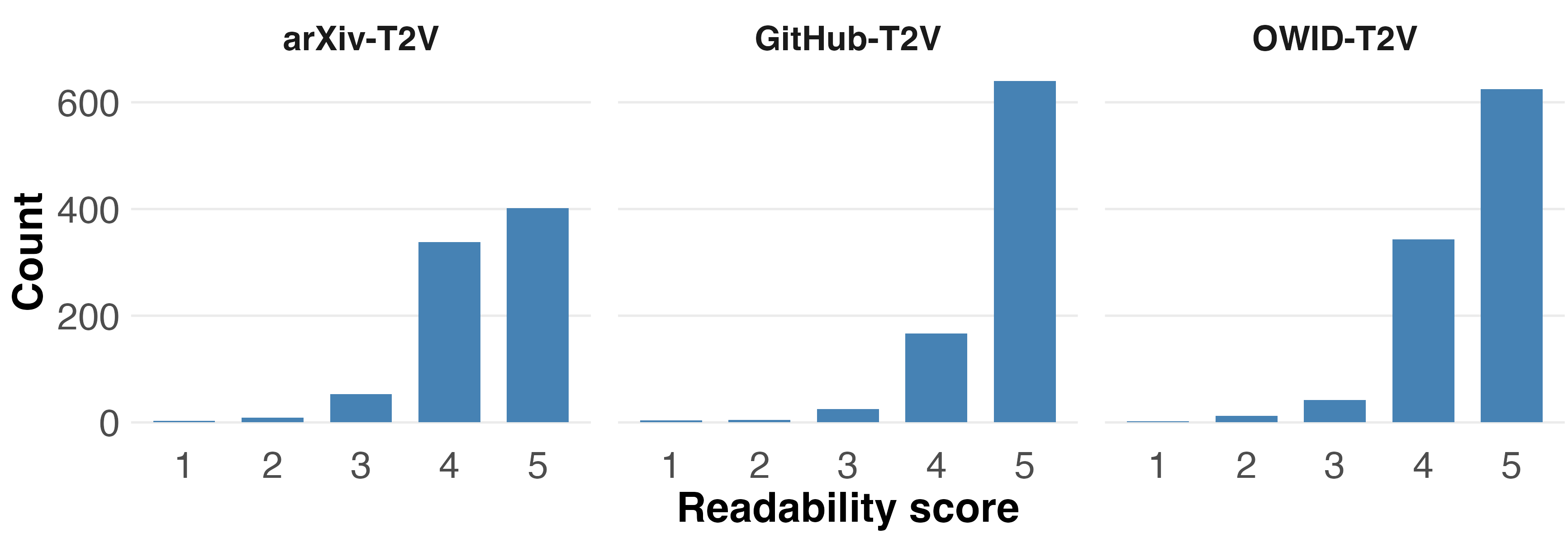}
    \caption{Distribution of readability scores for visualisations produced in run 1 for each dataset-agent combination (GitHub-T2V, arXiv-T2V, OWID-T2V; Co-GPT-5)}
    \label{fig:readability_score_hist}
\end{figure}

Figure~\ref {fig:readability_score_hist} shows the distribution of the readability scores (1-5) of these same generated visualisations. 
Across both datasets and models, readability scores are skewed to the high end of the scale. 
However, the arXiv-T2V scores are more dispersed between 4 and 5, while the GitHub-T2V and OWID-T2V scores are more concentrated at 5. 

\begin{figure}[t!]
    \centering
    \includegraphics[width=\linewidth]{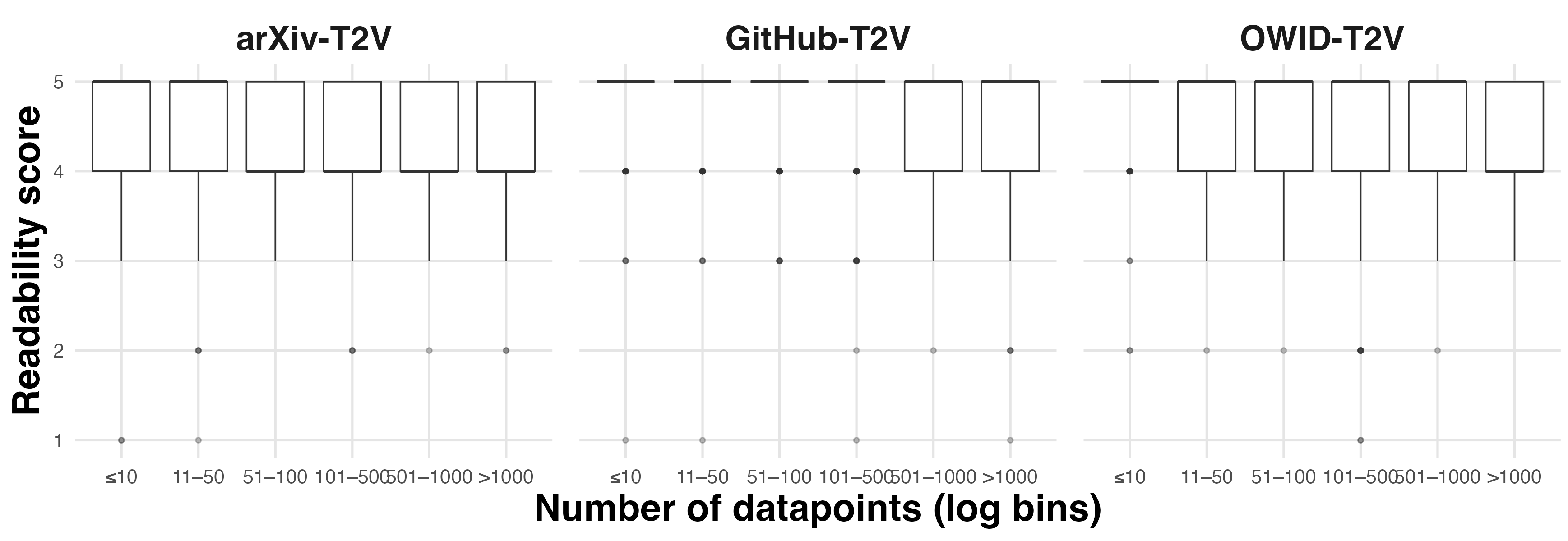}
    \caption{Readability score across logged bins of number of data points in visualisations produced in run 1 for each dataset-agent combination (GitHub-T2V, arXiv-T2V, OWID-T2V; Co-GPT-5)}
    \label{fig:readability_num_dtps_box}
\end{figure}

The number of data points in a visualisation was not meaningfully associated with the readability score for any of the runs investigated. 
For the arXiv-T2V dataset using human NL requests Co-GPT-5 agent, we found no significant Pearson coefficient, suggesting there was no linear association between request length and similarity score ($r$ = -0.039, $p$ = 0.269). 
The Spearman coefficient, while significant, was close to zero ($\rho$ = -0.153, $p$ = 1.38e-05), suggestive of only a weak monotonic association. 
This pattern was replicated across the other combinations of datasets and agents: Pearson coefficients were non-significant, and where Spearman correlations were significant, they were close to zero and not consistent in direction. 
There was thus no clear or consistent relationship between the number of data points in a visualisation and the readability score. 
Readability score across logged bins of the number of data points in a visualisation for each dataset-agent combination investigated is presented in Figure~\ref{fig:readability_num_dtps_box}.

\begin{figure}[t]
\centering
\resizebox{0.35\textwidth}{!}{
\begin{tikzpicture}
{'xy-axis', 'other', 'legend', 'grid', 'title-subtitle', 'annotation', 'figure-format', 'data-plotting'}
\fill[gray!10!white] (0.0000,0.0000) rectangle (2.8000,0.3950);
\node[anchor=south west, inner sep=2, font=\fontsize{8}{8}\selectfont] at (0,0.0000) {multiple-plots};
\fill[gray!10!white] (0.0000,0.3950) rectangle (2.8000,0.7900);
\node[anchor=south west, inner sep=2, font=\fontsize{8}{8}\selectfont] at (0,0.3950) {margin-padding};
\fill[gray!10!white] (0.0000,0.7900) rectangle (2.8000,1.1850);
\node[anchor=south west, inner sep=2, font=\fontsize{8}{8}\selectfont] at (0,0.7900) {background};
\fill[gray!10!white] (0.0000,1.1850) rectangle (2.8000,1.5800);
\node[anchor=south west, inner sep=2, font=\fontsize{8}{8}\selectfont] at (0,1.1850) {bounding-box-border};
\fill[cyan!10!white] (0.0000,1.5800) rectangle (2.8000,1.9750);
\node[anchor=south west, inner sep=2, font=\fontsize{8}{8}\selectfont] at (0,1.5800) {saving-format};
\fill[cyan!10!white] (0.0000,1.9750) rectangle (2.8000,2.3700);
\node[anchor=south west, inner sep=2, font=\fontsize{8}{8}\selectfont] at (0,1.9750) {dpi};
\fill[cyan!10!white] (0.0000,2.3700) rectangle (2.8000,2.7650);
\node[anchor=south west, inner sep=2, font=\fontsize{8}{8}\selectfont] at (0,2.3700) {size};
\fill[brown!10!white] (0.0000,2.7650) rectangle (2.8000,3.1600);
\node[anchor=south west, inner sep=2, font=\fontsize{8}{8}\selectfont] at (0,2.7650) {grid-linewidth};
\fill[brown!10!white] (0.0000,3.1600) rectangle (2.8000,3.5550);
\node[anchor=south west, inner sep=2, font=\fontsize{8}{8}\selectfont] at (0,3.1600) {grid-linestyle};
\fill[brown!10!white] (0.0000,3.5550) rectangle (2.8000,3.9500);
\node[anchor=south west, inner sep=2, font=\fontsize{8}{8}\selectfont] at (0,3.5550) {grid-color};
\fill[brown!10!white] (0.0000,3.9500) rectangle (2.8000,4.3450);
\node[anchor=south west, inner sep=2, font=\fontsize{8}{8}\selectfont] at (0,3.9500) {grid-visible};
\fill[red!10!white] (0.0000,4.3450) rectangle (2.8000,4.7400);
\node[anchor=south west, inner sep=2, font=\fontsize{8}{8}\selectfont] at (0,4.3450) {legend-is-display};
\fill[red!10!white] (0.0000,4.7400) rectangle (2.8000,5.1350);
\node[anchor=south west, inner sep=2, font=\fontsize{8}{8}\selectfont] at (0,4.7400) {legend-labels-color};
\fill[red!10!white] (0.0000,5.1350) rectangle (2.8000,5.5300);
\node[anchor=south west, inner sep=2, font=\fontsize{8}{8}\selectfont] at (0,5.1350) {legend-labels};
\fill[red!10!white] (0.0000,5.5300) rectangle (2.8000,5.9250);
\node[anchor=south west, inner sep=2, font=\fontsize{8}{8}\selectfont] at (0,5.5300) {legend-position};
\fill[red!10!white] (0.0000,5.9250) rectangle (2.8000,6.3200);
\node[anchor=south west, inner sep=2, font=\fontsize{8}{8}\selectfont] at (0,5.9250) {legend-fontsize};
\fill[red!10!white] (0.0000,6.3200) rectangle (2.8000,6.7150);
\node[anchor=south west, inner sep=2, font=\fontsize{8}{8}\selectfont] at (0,6.3200) {legend-title};
\fill[black!10!white] (0.0000,6.7150) rectangle (2.8000,7.1100);
\node[anchor=south west, inner sep=2, font=\fontsize{8}{8}\selectfont] at (0,6.7150) {subtitle-fontsize};
\fill[black!10!white] (0.0000,7.1100) rectangle (2.8000,7.5050);
\node[anchor=south west, inner sep=2, font=\fontsize{8}{8}\selectfont] at (0,7.1100) {subtitle};
\fill[black!10!white] (0.0000,7.5050) rectangle (2.8000,7.9000);
\node[anchor=south west, inner sep=2, font=\fontsize{8}{8}\selectfont] at (0,7.5050) {title-position};
\fill[black!10!white] (0.0000,7.9000) rectangle (2.8000,8.2950);
\node[anchor=south west, inner sep=2, font=\fontsize{8}{8}\selectfont] at (0,7.9000) {title-color};
\fill[black!10!white] (0.0000,8.2950) rectangle (2.8000,8.6900);
\node[anchor=south west, inner sep=2, font=\fontsize{8}{8}\selectfont] at (0,8.2950) {title-fontsize};
\fill[black!10!white] (0.0000,8.6900) rectangle (2.8000,9.0850);
\node[anchor=south west, inner sep=2, font=\fontsize{8}{8}\selectfont] at (0,8.6900) {title};
\fill[magenta!10!white] (0.0000,9.0850) rectangle (2.8000,9.4800);
\node[anchor=south west, inner sep=2, font=\fontsize{8}{8}\selectfont] at (0,9.0850) {ann-font};
\fill[magenta!10!white] (0.0000,9.4800) rectangle (2.8000,9.8750);
\node[anchor=south west, inner sep=2, font=\fontsize{8}{8}\selectfont] at (0,9.4800) {ann-possition};
\fill[magenta!10!white] (0.0000,9.8750) rectangle (2.8000,10.2700);
\node[anchor=south west, inner sep=2, font=\fontsize{8}{8}\selectfont] at (0,9.8750) {ann-fontsize};
\fill[magenta!10!white] (0.0000,10.2700) rectangle (2.8000,10.6650);
\node[anchor=south west, inner sep=2, font=\fontsize{8}{8}\selectfont] at (0,10.2700) {ann-text-label};
\fill[blue!10!white] (0.0000,10.6650) rectangle (2.8000,11.0600);
\node[anchor=south west, inner sep=2, font=\fontsize{8}{8}\selectfont] at (0,10.6650) {errbar-visible};
\fill[blue!10!white] (0.0000,11.0600) rectangle (2.8000,11.4550);
\node[anchor=south west, inner sep=2, font=\fontsize{8}{8}\selectfont] at (0,11.0600) {errbar-color};
\fill[blue!10!white] (0.0000,11.4550) rectangle (2.8000,11.8500);
\node[anchor=south west, inner sep=2, font=\fontsize{8}{8}\selectfont] at (0,11.4550) {errbar-cap-thick};
\fill[blue!10!white] (0.0000,11.8500) rectangle (2.8000,12.2450);
\node[anchor=south west, inner sep=2, font=\fontsize{8}{8}\selectfont] at (0,11.8500) {errbar-cap-size};
\fill[blue!10!white] (0.0000,12.2450) rectangle (2.8000,12.6400);
\node[anchor=south west, inner sep=2, font=\fontsize{8}{8}\selectfont] at (0,12.2450) {pie-radius};
\fill[blue!10!white] (0.0000,12.6400) rectangle (2.8000,13.0350);
\node[anchor=south west, inner sep=2, font=\fontsize{8}{8}\selectfont] at (0,12.6400) {pie-precision-digits};
\fill[blue!10!white] (0.0000,13.0350) rectangle (2.8000,13.4300);
\node[anchor=south west, inner sep=2, font=\fontsize{8}{8}\selectfont] at (0,13.0350) {pie-percentage-distance};
\fill[blue!10!white] (0.0000,13.4300) rectangle (2.8000,13.8250);
\node[anchor=south west, inner sep=2, font=\fontsize{8}{8}\selectfont] at (0,13.4300) {pie-label-distance};
\fill[blue!10!white] (0.0000,13.8250) rectangle (2.8000,14.2200);
\node[anchor=south west, inner sep=2, font=\fontsize{8}{8}\selectfont] at (0,13.8250) {pie-explode};
\fill[blue!10!white] (0.0000,14.2200) rectangle (2.8000,14.6150);
\node[anchor=south west, inner sep=2, font=\fontsize{8}{8}\selectfont] at (0,14.2200) {hist-bins};
\fill[blue!10!white] (0.0000,14.6150) rectangle (2.8000,15.0100);
\node[anchor=south west, inner sep=2, font=\fontsize{8}{8}\selectfont] at (0,14.6150) {bar-data-stacking};
\fill[blue!10!white] (0.0000,15.0100) rectangle (2.8000,15.4050);
\node[anchor=south west, inner sep=2, font=\fontsize{8}{8}\selectfont] at (0,15.0100) {bar-thickness};
\fill[blue!10!white] (0.0000,15.4050) rectangle (2.8000,15.8000);
\node[anchor=south west, inner sep=2, font=\fontsize{8}{8}\selectfont] at (0,15.4050) {markerstyle};
\fill[blue!10!white] (0.0000,15.8000) rectangle (2.8000,16.1950);
\node[anchor=south west, inner sep=2, font=\fontsize{8}{8}\selectfont] at (0,15.8000) {line-capstyle};
\fill[blue!10!white] (0.0000,16.1950) rectangle (2.8000,16.5900);
\node[anchor=south west, inner sep=2, font=\fontsize{8}{8}\selectfont] at (0,16.1950) {linestyle};
\fill[blue!10!white] (0.0000,16.5900) rectangle (2.8000,16.9850);
\node[anchor=south west, inner sep=2, font=\fontsize{8}{8}\selectfont] at (0,16.5900) {markersize};
\fill[blue!10!white] (0.0000,16.9850) rectangle (2.8000,17.3800);
\node[anchor=south west, inner sep=2, font=\fontsize{8}{8}\selectfont] at (0,16.9850) {linewidth};
\fill[blue!10!white] (0.0000,17.3800) rectangle (2.8000,17.7750);
\node[anchor=south west, inner sep=2, font=\fontsize{8}{8}\selectfont] at (0,17.3800) {opacity};
\fill[blue!10!white] (0.0000,17.7750) rectangle (2.8000,18.1700);
\node[anchor=south west, inner sep=2, font=\fontsize{8}{8}\selectfont] at (0,17.7750) {edge-color};
\fill[blue!10!white] (0.0000,18.1700) rectangle (2.8000,18.5650);
\node[anchor=south west, inner sep=2, font=\fontsize{8}{8}\selectfont] at (0,18.1700) {filled-color};
\fill[yellow!10!white] (0.0000,18.5650) rectangle (2.8000,18.9600);
\node[anchor=south west, inner sep=2, font=\fontsize{8}{8}\selectfont] at (0,18.5650) {invert-x-y-axis};
\fill[yellow!10!white] (0.0000,18.9600) rectangle (2.8000,19.3550);
\node[anchor=south west, inner sep=2, font=\fontsize{8}{8}\selectfont] at (0,18.9600) {x-y-scale-position};
\fill[yellow!10!white] (0.0000,19.3550) rectangle (2.8000,19.7500);
\node[anchor=south west, inner sep=2, font=\fontsize{8}{8}\selectfont] at (0,19.3550) {y-axis-ticks-visible};
\fill[yellow!10!white] (0.0000,19.7500) rectangle (2.8000,20.1450);
\node[anchor=south west, inner sep=2, font=\fontsize{8}{8}\selectfont] at (0,19.7500) {x-axis-ticks-visible};
\fill[yellow!10!white] (0.0000,20.1450) rectangle (2.8000,20.5400);
\node[anchor=south west, inner sep=2, font=\fontsize{8}{8}\selectfont] at (0,20.1450) {x-y-ticks-fontsize};
\fill[yellow!10!white] (0.0000,20.5400) rectangle (2.8000,20.9350);
\node[anchor=south west, inner sep=2, font=\fontsize{8}{8}\selectfont] at (0,20.5400) {x-y-scale};
\fill[yellow!10!white] (0.0000,20.9350) rectangle (2.8000,21.3300);
\node[anchor=south west, inner sep=2, font=\fontsize{8}{8}\selectfont] at (0,20.9350) {x-y-ticks-labels-rotation};
\fill[yellow!10!white] (0.0000,21.3300) rectangle (2.8000,21.7250);
\node[anchor=south west, inner sep=2, font=\fontsize{8}{8}\selectfont] at (0,21.3300) {x-y-ticks-labels-color};
\fill[yellow!10!white] (0.0000,21.7250) rectangle (2.8000,22.1200);
\node[anchor=south west, inner sep=2, font=\fontsize{8}{8}\selectfont] at (0,21.7250) {x-y-ticks-labels};
\fill[yellow!10!white] (0.0000,22.1200) rectangle (2.8000,22.5150);
\node[anchor=south west, inner sep=2, font=\fontsize{8}{8}\selectfont] at (0,22.1200) {x-y-lim};
\fill[yellow!10!white] (0.0000,22.5150) rectangle (2.8000,22.9100);
\node[anchor=south west, inner sep=2, font=\fontsize{8}{8}\selectfont] at (0,22.5150) {x-y-title-color};
\fill[yellow!10!white] (0.0000,22.9100) rectangle (2.8000,23.3050);
\node[anchor=south west, inner sep=2, font=\fontsize{8}{8}\selectfont] at (0,22.9100) {x-y-title-fontsize};
\fill[yellow!10!white] (0.0000,23.3050) rectangle (2.8000,23.7000);
\node[anchor=south west, inner sep=2, font=\fontsize{8}{8}\selectfont] at (0,23.3050) {y-title};
\fill[yellow!10!white] (0.0000,23.7000) rectangle (2.8000,24.0950);
\node[anchor=south west, inner sep=2, font=\fontsize{8}{8}\selectfont] at (0,23.7000) {x-title};
\node[anchor=south west, inner sep=2, font=\fontsize{8}{8}\selectfont] at (0,24.0950) {\textbf{Attribute}};
\fill[green!84!white] (2.8000,0.0000) rectangle (3.5500,0.3950);
\node[anchor=south west, inner sep=2, font=\fontsize{8}{8}\selectfont] at (2.9000,0.0000) {7.4};
\fill[green!35!white] (2.8000,0.3950) rectangle (3.5500,0.7900);
\node[anchor=south west, inner sep=2, font=\fontsize{8}{8}\selectfont] at (2.9000,0.3950) {3.1};
\fill[green!0!white] (2.8000,0.7900) rectangle (3.5500,1.1850);
\node[anchor=south west, inner sep=2, font=\fontsize{8}{8}\selectfont] at (2.9000,0.7900) {0.1};
\fill[green!0!white] (2.8000,1.1850) rectangle (3.5500,1.5800);
\node[anchor=south west, inner sep=2, font=\fontsize{8}{8}\selectfont] at (2.9000,1.1850) {0.0};
\fill[green!1!white] (2.8000,1.5800) rectangle (3.5500,1.9750);
\node[anchor=south west, inner sep=2, font=\fontsize{8}{8}\selectfont] at (2.9000,1.5800) {0.2};
\fill[green!10!white] (2.8000,1.9750) rectangle (3.5500,2.3700);
\node[anchor=south west, inner sep=2, font=\fontsize{8}{8}\selectfont] at (2.9000,1.9750) {0.9};
\fill[green!97!white] (2.8000,2.3700) rectangle (3.5500,2.7650);
\node[anchor=south west, inner sep=2, font=\fontsize{8}{8}\selectfont] at (2.9000,2.3700) {8.6};
\fill[green!0!white] (2.8000,2.7650) rectangle (3.5500,3.1600);
\node[anchor=south west, inner sep=2, font=\fontsize{8}{8}\selectfont] at (2.9000,2.7650) {0.1};
\fill[green!1!white] (2.8000,3.1600) rectangle (3.5500,3.5550);
\node[anchor=south west, inner sep=2, font=\fontsize{8}{8}\selectfont] at (2.9000,3.1600) {0.2};
\fill[green!1!white] (2.8000,3.5550) rectangle (3.5500,3.9500);
\node[anchor=south west, inner sep=2, font=\fontsize{8}{8}\selectfont] at (2.9000,3.5550) {0.1};
\fill[green!7!white] (2.8000,3.9500) rectangle (3.5500,4.3450);
\node[anchor=south west, inner sep=2, font=\fontsize{8}{8}\selectfont] at (2.9000,3.9500) {0.7};
\fill[green!59!white] (2.8000,4.3450) rectangle (3.5500,4.7400);
\node[anchor=south west, inner sep=2, font=\fontsize{8}{8}\selectfont] at (2.9000,4.3450) {5.2};
\fill[green!0!white] (2.8000,4.7400) rectangle (3.5500,5.1350);
\node[anchor=south west, inner sep=2, font=\fontsize{8}{8}\selectfont] at (2.9000,4.7400) {0.0};
\fill[green!93!white] (2.8000,5.1350) rectangle (3.5500,5.5300);
\node[anchor=south west, inner sep=2, font=\fontsize{8}{8}\selectfont] at (2.9000,5.1350) {8.2};
\fill[green!26!white] (2.8000,5.5300) rectangle (3.5500,5.9250);
\node[anchor=south west, inner sep=2, font=\fontsize{8}{8}\selectfont] at (2.9000,5.5300) {2.3};
\fill[green!5!white] (2.8000,5.9250) rectangle (3.5500,6.3200);
\node[anchor=south west, inner sep=2, font=\fontsize{8}{8}\selectfont] at (2.9000,5.9250) {0.5};
\fill[green!1!white] (2.8000,6.3200) rectangle (3.5500,6.7150);
\node[anchor=south west, inner sep=2, font=\fontsize{8}{8}\selectfont] at (2.9000,6.3200) {0.1};
\fill[green!2!white] (2.8000,6.7150) rectangle (3.5500,7.1100);
\node[anchor=south west, inner sep=2, font=\fontsize{8}{8}\selectfont] at (2.9000,6.7150) {0.2};
\fill[green!5!white] (2.8000,7.1100) rectangle (3.5500,7.5050);
\node[anchor=south west, inner sep=2, font=\fontsize{8}{8}\selectfont] at (2.9000,7.1100) {0.5};
\fill[green!0!white] (2.8000,7.5050) rectangle (3.5500,7.9000);
\node[anchor=south west, inner sep=2, font=\fontsize{8}{8}\selectfont] at (2.9000,7.5050) {0.1};
\fill[green!0!white] (2.8000,7.9000) rectangle (3.5500,8.2950);
\node[anchor=south west, inner sep=2, font=\fontsize{8}{8}\selectfont] at (2.9000,7.9000) {0.0};
\fill[green!14!white] (2.8000,8.2950) rectangle (3.5500,8.6900);
\node[anchor=south west, inner sep=2, font=\fontsize{8}{8}\selectfont] at (2.9000,8.2950) {1.2};
\fill[green!100!white] (2.8000,8.6900) rectangle (3.5500,9.0850);
\node[anchor=south west, inner sep=2, font=\fontsize{8}{8}\selectfont] at (2.9000,8.6900) {8.8};
\fill[green!1!white] (2.8000,9.0850) rectangle (3.5500,9.4800);
\node[anchor=south west, inner sep=2, font=\fontsize{8}{8}\selectfont] at (2.9000,9.0850) {0.1};
\fill[green!10!white] (2.8000,9.4800) rectangle (3.5500,9.8750);
\node[anchor=south west, inner sep=2, font=\fontsize{8}{8}\selectfont] at (2.9000,9.4800) {0.9};
\fill[green!8!white] (2.8000,9.8750) rectangle (3.5500,10.2700);
\node[anchor=south west, inner sep=2, font=\fontsize{8}{8}\selectfont] at (2.9000,9.8750) {0.8};
\fill[green!5!white] (2.8000,10.2700) rectangle (3.5500,10.6650);
\node[anchor=south west, inner sep=2, font=\fontsize{8}{8}\selectfont] at (2.9000,10.2700) {0.5};
\fill[green!4!white] (2.8000,10.6650) rectangle (3.5500,11.0600);
\node[anchor=south west, inner sep=2, font=\fontsize{8}{8}\selectfont] at (2.9000,10.6650) {0.4};
\fill[green!0!white] (2.8000,11.0600) rectangle (3.5500,11.4550);
\node[anchor=south west, inner sep=2, font=\fontsize{8}{8}\selectfont] at (2.9000,11.0600) {0.0};
\fill[green!0!white] (2.8000,11.4550) rectangle (3.5500,11.8500);
\node[anchor=south west, inner sep=2, font=\fontsize{8}{8}\selectfont] at (2.9000,11.4550) {0.0};
\fill[green!0!white] (2.8000,11.8500) rectangle (3.5500,12.2450);
\node[anchor=south west, inner sep=2, font=\fontsize{8}{8}\selectfont] at (2.9000,11.8500) {0.1};
\fill[green!0!white] (2.8000,12.2450) rectangle (3.5500,12.6400);
\node[anchor=south west, inner sep=2, font=\fontsize{8}{8}\selectfont] at (2.9000,12.2450) {0.0};
\fill[green!1!white] (2.8000,12.6400) rectangle (3.5500,13.0350);
\node[anchor=south west, inner sep=2, font=\fontsize{8}{8}\selectfont] at (2.9000,12.6400) {0.1};
\fill[green!0!white] (2.8000,13.0350) rectangle (3.5500,13.4300);
\node[anchor=south west, inner sep=2, font=\fontsize{8}{8}\selectfont] at (2.9000,13.0350) {0.0};
\fill[green!0!white] (2.8000,13.4300) rectangle (3.5500,13.8250);
\node[anchor=south west, inner sep=2, font=\fontsize{8}{8}\selectfont] at (2.9000,13.4300) {0.0};
\fill[green!0!white] (2.8000,13.8250) rectangle (3.5500,14.2200);
\node[anchor=south west, inner sep=2, font=\fontsize{8}{8}\selectfont] at (2.9000,13.8250) {0.1};
\fill[green!11!white] (2.8000,14.2200) rectangle (3.5500,14.6150);
\node[anchor=south west, inner sep=2, font=\fontsize{8}{8}\selectfont] at (2.9000,14.2200) {1.0};
\fill[green!0!white] (2.8000,14.6150) rectangle (3.5500,15.0100);
\node[anchor=south west, inner sep=2, font=\fontsize{8}{8}\selectfont] at (2.9000,14.6150) {0.0};
\fill[green!5!white] (2.8000,15.0100) rectangle (3.5500,15.4050);
\node[anchor=south west, inner sep=2, font=\fontsize{8}{8}\selectfont] at (2.9000,15.0100) {0.4};
\fill[green!16!white] (2.8000,15.4050) rectangle (3.5500,15.8000);
\node[anchor=south west, inner sep=2, font=\fontsize{8}{8}\selectfont] at (2.9000,15.4050) {1.4};
\fill[green!0!white] (2.8000,15.8000) rectangle (3.5500,16.1950);
\node[anchor=south west, inner sep=2, font=\fontsize{8}{8}\selectfont] at (2.9000,15.8000) {0.0};
\fill[green!11!white] (2.8000,16.1950) rectangle (3.5500,16.5900);
\node[anchor=south west, inner sep=2, font=\fontsize{8}{8}\selectfont] at (2.9000,16.1950) {1.0};
\fill[green!20!white] (2.8000,16.5900) rectangle (3.5500,16.9850);
\node[anchor=south west, inner sep=2, font=\fontsize{8}{8}\selectfont] at (2.9000,16.5900) {1.8};
\fill[green!28!white] (2.8000,16.9850) rectangle (3.5500,17.3800);
\node[anchor=south west, inner sep=2, font=\fontsize{8}{8}\selectfont] at (2.9000,16.9850) {2.5};
\fill[green!27!white] (2.8000,17.3800) rectangle (3.5500,17.7750);
\node[anchor=south west, inner sep=2, font=\fontsize{8}{8}\selectfont] at (2.9000,17.3800) {2.4};
\fill[green!7!white] (2.8000,17.7750) rectangle (3.5500,18.1700);
\node[anchor=south west, inner sep=2, font=\fontsize{8}{8}\selectfont] at (2.9000,17.7750) {0.7};
\fill[green!96!white] (2.8000,18.1700) rectangle (3.5500,18.5650);
\node[anchor=south west, inner sep=2, font=\fontsize{8}{8}\selectfont] at (2.9000,18.1700) {8.5};
\fill[green!0!white] (2.8000,18.5650) rectangle (3.5500,18.9600);
\node[anchor=south west, inner sep=2, font=\fontsize{8}{8}\selectfont] at (2.9000,18.5650) {0.1};
\fill[green!2!white] (2.8000,18.9600) rectangle (3.5500,19.3550);
\node[anchor=south west, inner sep=2, font=\fontsize{8}{8}\selectfont] at (2.9000,18.9600) {0.2};
\fill[green!4!white] (2.8000,19.3550) rectangle (3.5500,19.7500);
\node[anchor=south west, inner sep=2, font=\fontsize{8}{8}\selectfont] at (2.9000,19.3550) {0.4};
\fill[green!8!white] (2.8000,19.7500) rectangle (3.5500,20.1450);
\node[anchor=south west, inner sep=2, font=\fontsize{8}{8}\selectfont] at (2.9000,19.7500) {0.7};
\fill[green!9!white] (2.8000,20.1450) rectangle (3.5500,20.5400);
\node[anchor=south west, inner sep=2, font=\fontsize{8}{8}\selectfont] at (2.9000,20.1450) {0.9};
\fill[green!4!white] (2.8000,20.5400) rectangle (3.5500,20.9350);
\node[anchor=south west, inner sep=2, font=\fontsize{8}{8}\selectfont] at (2.9000,20.5400) {0.4};
\fill[green!7!white] (2.8000,20.9350) rectangle (3.5500,21.3300);
\node[anchor=south west, inner sep=2, font=\fontsize{8}{8}\selectfont] at (2.9000,20.9350) {0.7};
\fill[green!0!white] (2.8000,21.3300) rectangle (3.5500,21.7250);
\node[anchor=south west, inner sep=2, font=\fontsize{8}{8}\selectfont] at (2.9000,21.3300) {0.1};
\fill[green!35!white] (2.8000,21.7250) rectangle (3.5500,22.1200);
\node[anchor=south west, inner sep=2, font=\fontsize{8}{8}\selectfont] at (2.9000,21.7250) {3.1};
\fill[green!27!white] (2.8000,22.1200) rectangle (3.5500,22.5150);
\node[anchor=south west, inner sep=2, font=\fontsize{8}{8}\selectfont] at (2.9000,22.1200) {2.4};
\fill[green!1!white] (2.8000,22.5150) rectangle (3.5500,22.9100);
\node[anchor=south west, inner sep=2, font=\fontsize{8}{8}\selectfont] at (2.9000,22.5150) {0.2};
\fill[green!30!white] (2.8000,22.9100) rectangle (3.5500,23.3050);
\node[anchor=south west, inner sep=2, font=\fontsize{8}{8}\selectfont] at (2.9000,22.9100) {2.7};
\fill[green!98!white] (2.8000,23.3050) rectangle (3.5500,23.7000);
\node[anchor=south west, inner sep=2, font=\fontsize{8}{8}\selectfont] at (2.9000,23.3050) {8.6};
\fill[green!97!white] (2.8000,23.7000) rectangle (3.5500,24.0950);
\node[anchor=south west, inner sep=2, font=\fontsize{8}{8}\selectfont] at (2.9000,23.7000) {8.5};
\node[anchor=south west, inner sep=2, rotate=20, yshift=0cm, xshift=0cm ,font=\fontsize{8}{8}\selectfont] at (2.8000,24.0950) {\textbf{Matplotlib-nb}};
% +++++++++++++++++++++++++++++++++++++++++++++++++++++++++++++++
\fill[green!57!white] (3.5500,0.0000) rectangle (4.3000,0.3950);
\node[anchor=south west, inner sep=2, font=\fontsize{8}{8}\selectfont] at (3.6500,0.0000) {7.4};
\fill[green!11!white] (3.5500,0.3950) rectangle (4.3000,0.7900);
\node[anchor=south west, inner sep=2, font=\fontsize{8}{8}\selectfont] at (3.6500,0.3950) {1.5};
\fill[green!0!white] (3.5500,0.7900) rectangle (4.3000,1.1850);
\node[anchor=south west, inner sep=2, font=\fontsize{8}{8}\selectfont] at (3.6500,0.7900) {0.0};
\fill[green!0!white] (3.5500,1.1850) rectangle (4.3000,1.5800);
\node[anchor=south west, inner sep=2, font=\fontsize{8}{8}\selectfont] at (3.6500,1.1850) {0.0};
\fill[green!0!white] (3.5500,1.5800) rectangle (4.3000,1.9750);
\node[anchor=south west, inner sep=2, font=\fontsize{8}{8}\selectfont] at (3.6500,1.5800) {0.0};
\fill[green!1!white] (3.5500,1.9750) rectangle (4.3000,2.3700);
\node[anchor=south west, inner sep=2, font=\fontsize{8}{8}\selectfont] at (3.6500,1.9750) {0.2};
\fill[green!57!white] (3.5500,2.3700) rectangle (4.3000,2.7650);
\node[anchor=south west, inner sep=2, font=\fontsize{8}{8}\selectfont] at (3.6500,2.3700) {7.4};
\fill[green!0!white] (3.5500,2.7650) rectangle (4.3000,3.1600);
\node[anchor=south west, inner sep=2, font=\fontsize{8}{8}\selectfont] at (3.6500,2.7650) {0.0};
\fill[green!0!white] (3.5500,3.1600) rectangle (4.3000,3.5550);
\node[anchor=south west, inner sep=2, font=\fontsize{8}{8}\selectfont] at (3.6500,3.1600) {0.0};
\fill[green!0!white] (3.5500,3.5550) rectangle (4.3000,3.9500);
\node[anchor=south west, inner sep=2, font=\fontsize{8}{8}\selectfont] at (3.6500,3.5550) {0.0};
\fill[green!0!white] (3.5500,3.9500) rectangle (4.3000,4.3450);
\node[anchor=south west, inner sep=2, font=\fontsize{8}{8}\selectfont] at (3.6500,3.9500) {0.1};
\fill[green!30!white] (3.5500,4.3450) rectangle (4.3000,4.7400);
\node[anchor=south west, inner sep=2, font=\fontsize{8}{8}\selectfont] at (3.6500,4.3450) {3.9};
\fill[green!0!white] (3.5500,4.7400) rectangle (4.3000,5.1350);
\node[anchor=south west, inner sep=2, font=\fontsize{8}{8}\selectfont] at (3.6500,4.7400) {0.0};
\fill[green!49!white] (3.5500,5.1350) rectangle (4.3000,5.5300);
\node[anchor=south west, inner sep=2, font=\fontsize{8}{8}\selectfont] at (3.6500,5.1350) {6.4};
\fill[green!12!white] (3.5500,5.5300) rectangle (4.3000,5.9250);
\node[anchor=south west, inner sep=2, font=\fontsize{8}{8}\selectfont] at (3.6500,5.5300) {1.7};
\fill[green!0!white] (3.5500,5.9250) rectangle (4.3000,6.3200);
\node[anchor=south west, inner sep=2, font=\fontsize{8}{8}\selectfont] at (3.6500,5.9250) {0.1};
\fill[green!3!white] (3.5500,6.3200) rectangle (4.3000,6.7150);
\node[anchor=south west, inner sep=2, font=\fontsize{8}{8}\selectfont] at (3.6500,6.3200) {0.4};
\fill[green!0!white] (3.5500,6.7150) rectangle (4.3000,7.1100);
\node[anchor=south west, inner sep=2, font=\fontsize{8}{8}\selectfont] at (3.6500,6.7150) {0.1};
\fill[green!3!white] (3.5500,7.1100) rectangle (4.3000,7.5050);
\node[anchor=south west, inner sep=2, font=\fontsize{8}{8}\selectfont] at (3.6500,7.1100) {0.4};
\fill[green!0!white] (3.5500,7.5050) rectangle (4.3000,7.9000);
\node[anchor=south west, inner sep=2, font=\fontsize{8}{8}\selectfont] at (3.6500,7.5050) {0.0};
\fill[green!0!white] (3.5500,7.9000) rectangle (4.3000,8.2950);
\node[anchor=south west, inner sep=2, font=\fontsize{8}{8}\selectfont] at (3.6500,7.9000) {0.0};
\fill[green!3!white] (3.5500,8.2950) rectangle (4.3000,8.6900);
\node[anchor=south west, inner sep=2, font=\fontsize{8}{8}\selectfont] at (3.6500,8.2950) {0.4};
\fill[green!64!white] (3.5500,8.6900) rectangle (4.3000,9.0850);
\node[anchor=south west, inner sep=2, font=\fontsize{8}{8}\selectfont] at (3.6500,8.6900) {8.3};
\fill[green!0!white] (3.5500,9.0850) rectangle (4.3000,9.4800);
\node[anchor=south west, inner sep=2, font=\fontsize{8}{8}\selectfont] at (3.6500,9.0850) {0.0};
\fill[green!0!white] (3.5500,9.4800) rectangle (4.3000,9.8750);
\node[anchor=south west, inner sep=2, font=\fontsize{8}{8}\selectfont] at (3.6500,9.4800) {0.0};
\fill[green!21!white] (3.5500,9.8750) rectangle (4.3000,10.2700);
\node[anchor=south west, inner sep=2, font=\fontsize{8}{8}\selectfont] at (3.6500,9.8750) {2.8};
\fill[green!2!white] (3.5500,10.2700) rectangle (4.3000,10.6650);
\node[anchor=south west, inner sep=2, font=\fontsize{8}{8}\selectfont] at (3.6500,10.2700) {0.3};
\fill[green!1!white] (3.5500,10.6650) rectangle (4.3000,11.0600);
\node[anchor=south west, inner sep=2, font=\fontsize{8}{8}\selectfont] at (3.6500,10.6650) {0.3};
\fill[green!0!white] (3.5500,11.0600) rectangle (4.3000,11.4550);
\node[anchor=south west, inner sep=2, font=\fontsize{8}{8}\selectfont] at (3.6500,11.0600) {0.0};
\fill[green!0!white] (3.5500,11.4550) rectangle (4.3000,11.8500);
\node[anchor=south west, inner sep=2, font=\fontsize{8}{8}\selectfont] at (3.6500,11.4550) {0.0};
\fill[green!0!white] (3.5500,11.8500) rectangle (4.3000,12.2450);
\node[anchor=south west, inner sep=2, font=\fontsize{8}{8}\selectfont] at (3.6500,11.8500) {0.0};
\fill[green!0!white] (3.5500,12.2450) rectangle (4.3000,12.6400);
\node[anchor=south west, inner sep=2, font=\fontsize{8}{8}\selectfont] at (3.6500,12.2450) {0.0};
\fill[green!0!white] (3.5500,12.6400) rectangle (4.3000,13.0350);
\node[anchor=south west, inner sep=2, font=\fontsize{8}{8}\selectfont] at (3.6500,12.6400) {0.1};
\fill[green!0!white] (3.5500,13.0350) rectangle (4.3000,13.4300);
\node[anchor=south west, inner sep=2, font=\fontsize{8}{8}\selectfont] at (3.6500,13.0350) {0.0};
\fill[green!0!white] (3.5500,13.4300) rectangle (4.3000,13.8250);
\node[anchor=south west, inner sep=2, font=\fontsize{8}{8}\selectfont] at (3.6500,13.4300) {0.0};
\fill[green!0!white] (3.5500,13.8250) rectangle (4.3000,14.2200);
\node[anchor=south west, inner sep=2, font=\fontsize{8}{8}\selectfont] at (3.6500,13.8250) {0.0};
\fill[green!10!white] (3.5500,14.2200) rectangle (4.3000,14.6150);
\node[anchor=south west, inner sep=2, font=\fontsize{8}{8}\selectfont] at (3.6500,14.2200) {1.4};
\fill[green!0!white] (3.5500,14.6150) rectangle (4.3000,15.0100);
\node[anchor=south west, inner sep=2, font=\fontsize{8}{8}\selectfont] at (3.6500,14.6150) {0.0};
\fill[green!2!white] (3.5500,15.0100) rectangle (4.3000,15.4050);
\node[anchor=south west, inner sep=2, font=\fontsize{8}{8}\selectfont] at (3.6500,15.0100) {0.3};
\fill[green!21!white] (3.5500,15.4050) rectangle (4.3000,15.8000);
\node[anchor=south west, inner sep=2, font=\fontsize{8}{8}\selectfont] at (3.6500,15.4050) {2.7};
\fill[green!0!white] (3.5500,15.8000) rectangle (4.3000,16.1950);
\node[anchor=south west, inner sep=2, font=\fontsize{8}{8}\selectfont] at (3.6500,15.8000) {0.0};
\fill[green!27!white] (3.5500,16.1950) rectangle (4.3000,16.5900);
\node[anchor=south west, inner sep=2, font=\fontsize{8}{8}\selectfont] at (3.6500,16.1950) {3.5};
\fill[green!19!white] (3.5500,16.5900) rectangle (4.3000,16.9850);
\node[anchor=south west, inner sep=2, font=\fontsize{8}{8}\selectfont] at (3.6500,16.5900) {2.5};
\fill[green!48!white] (3.5500,16.9850) rectangle (4.3000,17.3800);
\node[anchor=south west, inner sep=2, font=\fontsize{8}{8}\selectfont] at (3.6500,16.9850) {6.2};
\fill[green!26!white] (3.5500,17.3800) rectangle (4.3000,17.7750);
\node[anchor=south west, inner sep=2, font=\fontsize{8}{8}\selectfont] at (3.6500,17.3800) {3.4};
\fill[green!5!white] (3.5500,17.7750) rectangle (4.3000,18.1700);
\node[anchor=south west, inner sep=2, font=\fontsize{8}{8}\selectfont] at (3.6500,17.7750) {0.7};
\fill[green!100!white] (3.5500,18.1700) rectangle (4.3000,18.5650);
\node[anchor=south west, inner sep=2, font=\fontsize{8}{8}\selectfont] at (3.6500,18.1700) {12.9};
\fill[green!0!white] (3.5500,18.5650) rectangle (4.3000,18.9600);
\node[anchor=south west, inner sep=2, font=\fontsize{8}{8}\selectfont] at (3.6500,18.5650) {0.0};
\fill[green!0!white] (3.5500,18.9600) rectangle (4.3000,19.3550);
\node[anchor=south west, inner sep=2, font=\fontsize{8}{8}\selectfont] at (3.6500,18.9600) {0.0};
\fill[green!1!white] (3.5500,19.3550) rectangle (4.3000,19.7500);
\node[anchor=south west, inner sep=2, font=\fontsize{8}{8}\selectfont] at (3.6500,19.3550) {0.2};
\fill[green!7!white] (3.5500,19.7500) rectangle (4.3000,20.1450);
\node[anchor=south west, inner sep=2, font=\fontsize{8}{8}\selectfont] at (3.6500,19.7500) {1.0};
\fill[green!1!white] (3.5500,20.1450) rectangle (4.3000,20.5400);
\node[anchor=south west, inner sep=2, font=\fontsize{8}{8}\selectfont] at (3.6500,20.1450) {0.1};
\fill[green!1!white] (3.5500,20.5400) rectangle (4.3000,20.9350);
\node[anchor=south west, inner sep=2, font=\fontsize{8}{8}\selectfont] at (3.6500,20.5400) {0.1};
\fill[green!3!white] (3.5500,20.9350) rectangle (4.3000,21.3300);
\node[anchor=south west, inner sep=2, font=\fontsize{8}{8}\selectfont] at (3.6500,20.9350) {0.4};
\fill[green!0!white] (3.5500,21.3300) rectangle (4.3000,21.7250);
\node[anchor=south west, inner sep=2, font=\fontsize{8}{8}\selectfont] at (3.6500,21.3300) {0.0};
\fill[green!12!white] (3.5500,21.7250) rectangle (4.3000,22.1200);
\node[anchor=south west, inner sep=2, font=\fontsize{8}{8}\selectfont] at (3.6500,21.7250) {1.6};
\fill[green!6!white] (3.5500,22.1200) rectangle (4.3000,22.5150);
\node[anchor=south west, inner sep=2, font=\fontsize{8}{8}\selectfont] at (3.6500,22.1200) {0.9};
\fill[green!0!white] (3.5500,22.5150) rectangle (4.3000,22.9100);
\node[anchor=south west, inner sep=2, font=\fontsize{8}{8}\selectfont] at (3.6500,22.5150) {0.0};
\fill[green!9!white] (3.5500,22.9100) rectangle (4.3000,23.3050);
\node[anchor=south west, inner sep=2, font=\fontsize{8}{8}\selectfont] at (3.6500,22.9100) {1.2};
\fill[green!69!white] (3.5500,23.3050) rectangle (4.3000,23.7000);
\node[anchor=south west, inner sep=2, font=\fontsize{8}{8}\selectfont] at (3.6500,23.3050) {9.0};
\fill[green!77!white] (3.5500,23.7000) rectangle (4.3000,24.0950);
\node[anchor=south west, inner sep=2, font=\fontsize{8}{8}\selectfont] at (3.6500,23.7000) {10.0};
\node[anchor=south west, inner sep=2, rotate=20, yshift=0cm, xshift=0cm ,font=\fontsize{8}{8}\selectfont] at (3.5500,24.0950) {\textbf{\textcolor{red}{GitHub-T2V}}};
% +++++++++++++++++++++++++++++++++++++++++++++++++++++++++++++++
\fill[green!38!white] (4.3000,0.0000) rectangle (5.0500,0.3950);
\node[anchor=south west, inner sep=2, font=\fontsize{8}{8}\selectfont] at (4.4000,0.0000) {3.5};
\fill[green!80!white] (4.3000,0.3950) rectangle (5.0500,0.7900);
\node[anchor=south west, inner sep=2, font=\fontsize{8}{8}\selectfont] at (4.4000,0.3950) {7.4};
\fill[green!1!white] (4.3000,0.7900) rectangle (5.0500,1.1850);
\node[anchor=south west, inner sep=2, font=\fontsize{8}{8}\selectfont] at (4.4000,0.7900) {0.1};
\fill[green!0!white] (4.3000,1.1850) rectangle (5.0500,1.5800);
\node[anchor=south west, inner sep=2, font=\fontsize{8}{8}\selectfont] at (4.4000,1.1850) {0.0};
\fill[green!4!white] (4.3000,1.5800) rectangle (5.0500,1.9750);
\node[anchor=south west, inner sep=2, font=\fontsize{8}{8}\selectfont] at (4.4000,1.5800) {0.4};
\fill[green!32!white] (4.3000,1.9750) rectangle (5.0500,2.3700);
\node[anchor=south west, inner sep=2, font=\fontsize{8}{8}\selectfont] at (4.4000,1.9750) {3.0};
\fill[green!49!white] (4.3000,2.3700) rectangle (5.0500,2.7650);
\node[anchor=south west, inner sep=2, font=\fontsize{8}{8}\selectfont] at (4.4000,2.3700) {4.5};
\fill[green!5!white] (4.3000,2.7650) rectangle (5.0500,3.1600);
\node[anchor=south west, inner sep=2, font=\fontsize{8}{8}\selectfont] at (4.4000,2.7650) {0.5};
\fill[green!10!white] (4.3000,3.1600) rectangle (5.0500,3.5550);
\node[anchor=south west, inner sep=2, font=\fontsize{8}{8}\selectfont] at (4.4000,3.1600) {1.0};
\fill[green!3!white] (4.3000,3.5550) rectangle (5.0500,3.9500);
\node[anchor=south west, inner sep=2, font=\fontsize{8}{8}\selectfont] at (4.4000,3.5550) {0.3};
\fill[green!12!white] (4.3000,3.9500) rectangle (5.0500,4.3450);
\node[anchor=south west, inner sep=2, font=\fontsize{8}{8}\selectfont] at (4.4000,3.9500) {1.1};
\fill[green!40!white] (4.3000,4.3450) rectangle (5.0500,4.7400);
\node[anchor=south west, inner sep=2, font=\fontsize{8}{8}\selectfont] at (4.4000,4.3450) {3.7};
\fill[green!0!white] (4.3000,4.7400) rectangle (5.0500,5.1350);
\node[anchor=south west, inner sep=2, font=\fontsize{8}{8}\selectfont] at (4.4000,4.7400) {0.0};
\fill[green!79!white] (4.3000,5.1350) rectangle (5.0500,5.5300);
\node[anchor=south west, inner sep=2, font=\fontsize{8}{8}\selectfont] at (4.4000,5.1350) {7.2};
\fill[green!39!white] (4.3000,5.5300) rectangle (5.0500,5.9250);
\node[anchor=south west, inner sep=2, font=\fontsize{8}{8}\selectfont] at (4.4000,5.5300) {3.6};
\fill[green!17!white] (4.3000,5.9250) rectangle (5.0500,6.3200);
\node[anchor=south west, inner sep=2, font=\fontsize{8}{8}\selectfont] at (4.4000,5.9250) {1.6};
\fill[green!2!white] (4.3000,6.3200) rectangle (5.0500,6.7150);
\node[anchor=south west, inner sep=2, font=\fontsize{8}{8}\selectfont] at (4.4000,6.3200) {0.2};
\fill[green!0!white] (4.3000,6.7150) rectangle (5.0500,7.1100);
\node[anchor=south west, inner sep=2, font=\fontsize{8}{8}\selectfont] at (4.4000,6.7150) {0.1};
\fill[green!1!white] (4.3000,7.1100) rectangle (5.0500,7.5050);
\node[anchor=south west, inner sep=2, font=\fontsize{8}{8}\selectfont] at (4.4000,7.1100) {0.1};
\fill[green!0!white] (4.3000,7.5050) rectangle (5.0500,7.9000);
\node[anchor=south west, inner sep=2, font=\fontsize{8}{8}\selectfont] at (4.4000,7.5050) {0.0};
\fill[green!0!white] (4.3000,7.9000) rectangle (5.0500,8.2950);
\node[anchor=south west, inner sep=2, font=\fontsize{8}{8}\selectfont] at (4.4000,7.9000) {0.0};
\fill[green!8!white] (4.3000,8.2950) rectangle (5.0500,8.6900);
\node[anchor=south west, inner sep=2, font=\fontsize{8}{8}\selectfont] at (4.4000,8.2950) {0.7};
\fill[green!19!white] (4.3000,8.6900) rectangle (5.0500,9.0850);
\node[anchor=south west, inner sep=2, font=\fontsize{8}{8}\selectfont] at (4.4000,8.6900) {1.8};
\fill[green!3!white] (4.3000,9.0850) rectangle (5.0500,9.4800);
\node[anchor=south west, inner sep=2, font=\fontsize{8}{8}\selectfont] at (4.4000,9.0850) {0.3};
\fill[green!37!white] (4.3000,9.4800) rectangle (5.0500,9.8750);
\node[anchor=south west, inner sep=2, font=\fontsize{8}{8}\selectfont] at (4.4000,9.4800) {3.4};
\fill[green!21!white] (4.3000,9.8750) rectangle (5.0500,10.2700);
\node[anchor=south west, inner sep=2, font=\fontsize{8}{8}\selectfont] at (4.4000,9.8750) {2.0};
\fill[green!10!white] (4.3000,10.2700) rectangle (5.0500,10.6650);
\node[anchor=south west, inner sep=2, font=\fontsize{8}{8}\selectfont] at (4.4000,10.2700) {0.9};
\fill[green!4!white] (4.3000,10.6650) rectangle (5.0500,11.0600);
\node[anchor=south west, inner sep=2, font=\fontsize{8}{8}\selectfont] at (4.4000,10.6650) {0.4};
\fill[green!0!white] (4.3000,11.0600) rectangle (5.0500,11.4550);
\node[anchor=south west, inner sep=2, font=\fontsize{8}{8}\selectfont] at (4.4000,11.0600) {0.1};
\fill[green!0!white] (4.3000,11.4550) rectangle (5.0500,11.8500);
\node[anchor=south west, inner sep=2, font=\fontsize{8}{8}\selectfont] at (4.4000,11.4550) {0.0};
\fill[green!1!white] (4.3000,11.8500) rectangle (5.0500,12.2450);
\node[anchor=south west, inner sep=2, font=\fontsize{8}{8}\selectfont] at (4.4000,11.8500) {0.2};
\fill[green!0!white] (4.3000,12.2450) rectangle (5.0500,12.6400);
\node[anchor=south west, inner sep=2, font=\fontsize{8}{8}\selectfont] at (4.4000,12.2450) {0.0};
\fill[green!0!white] (4.3000,12.6400) rectangle (5.0500,13.0350);
\node[anchor=south west, inner sep=2, font=\fontsize{8}{8}\selectfont] at (4.4000,12.6400) {0.0};
\fill[green!0!white] (4.3000,13.0350) rectangle (5.0500,13.4300);
\node[anchor=south west, inner sep=2, font=\fontsize{8}{8}\selectfont] at (4.4000,13.0350) {0.0};
\fill[green!0!white] (4.3000,13.4300) rectangle (5.0500,13.8250);
\node[anchor=south west, inner sep=2, font=\fontsize{8}{8}\selectfont] at (4.4000,13.4300) {0.0};
\fill[green!0!white] (4.3000,13.8250) rectangle (5.0500,14.2200);
\node[anchor=south west, inner sep=2, font=\fontsize{8}{8}\selectfont] at (4.4000,13.8250) {0.0};
\fill[green!1!white] (4.3000,14.2200) rectangle (5.0500,14.6150);
\node[anchor=south west, inner sep=2, font=\fontsize{8}{8}\selectfont] at (4.4000,14.2200) {0.1};
\fill[green!0!white] (4.3000,14.6150) rectangle (5.0500,15.0100);
\node[anchor=south west, inner sep=2, font=\fontsize{8}{8}\selectfont] at (4.4000,14.6150) {0.1};
\fill[green!21!white] (4.3000,15.0100) rectangle (5.0500,15.4050);
\node[anchor=south west, inner sep=2, font=\fontsize{8}{8}\selectfont] at (4.4000,15.0100) {1.9};
\fill[green!31!white] (4.3000,15.4050) rectangle (5.0500,15.8000);
\node[anchor=south west, inner sep=2, font=\fontsize{8}{8}\selectfont] at (4.4000,15.4050) {2.9};
\fill[green!0!white] (4.3000,15.8000) rectangle (5.0500,16.1950);
\node[anchor=south west, inner sep=2, font=\fontsize{8}{8}\selectfont] at (4.4000,15.8000) {0.0};
\fill[green!28!white] (4.3000,16.1950) rectangle (5.0500,16.5900);
\node[anchor=south west, inner sep=2, font=\fontsize{8}{8}\selectfont] at (4.4000,16.1950) {2.6};
\fill[green!20!white] (4.3000,16.5900) rectangle (5.0500,16.9850);
\node[anchor=south west, inner sep=2, font=\fontsize{8}{8}\selectfont] at (4.4000,16.5900) {1.9};
\fill[green!38!white] (4.3000,16.9850) rectangle (5.0500,17.3800);
\node[anchor=south west, inner sep=2, font=\fontsize{8}{8}\selectfont] at (4.4000,16.9850) {3.5};
\fill[green!27!white] (4.3000,17.3800) rectangle (5.0500,17.7750);
\node[anchor=south west, inner sep=2, font=\fontsize{8}{8}\selectfont] at (4.4000,17.3800) {2.5};
\fill[green!11!white] (4.3000,17.7750) rectangle (5.0500,18.1700);
\node[anchor=south west, inner sep=2, font=\fontsize{8}{8}\selectfont] at (4.4000,17.7750) {1.0};
\fill[green!100!white] (4.3000,18.1700) rectangle (5.0500,18.5650);
\node[anchor=south west, inner sep=2, font=\fontsize{8}{8}\selectfont] at (4.4000,18.1700) {9.1};
\fill[green!0!white] (4.3000,18.5650) rectangle (5.0500,18.9600);
\node[anchor=south west, inner sep=2, font=\fontsize{8}{8}\selectfont] at (4.4000,18.5650) {0.0};
\fill[green!5!white] (4.3000,18.9600) rectangle (5.0500,19.3550);
\node[anchor=south west, inner sep=2, font=\fontsize{8}{8}\selectfont] at (4.4000,18.9600) {0.5};
\fill[green!9!white] (4.3000,19.3550) rectangle (5.0500,19.7500);
\node[anchor=south west, inner sep=2, font=\fontsize{8}{8}\selectfont] at (4.4000,19.3550) {0.9};
\fill[green!20!white] (4.3000,19.7500) rectangle (5.0500,20.1450);
\node[anchor=south west, inner sep=2, font=\fontsize{8}{8}\selectfont] at (4.4000,19.7500) {1.9};
\fill[green!33!white] (4.3000,20.1450) rectangle (5.0500,20.5400);
\node[anchor=south west, inner sep=2, font=\fontsize{8}{8}\selectfont] at (4.4000,20.1450) {3.1};
\fill[green!2!white] (4.3000,20.5400) rectangle (5.0500,20.9350);
\node[anchor=south west, inner sep=2, font=\fontsize{8}{8}\selectfont] at (4.4000,20.5400) {0.2};
\fill[green!8!white] (4.3000,20.9350) rectangle (5.0500,21.3300);
\node[anchor=south west, inner sep=2, font=\fontsize{8}{8}\selectfont] at (4.4000,20.9350) {0.8};
\fill[green!3!white] (4.3000,21.3300) rectangle (5.0500,21.7250);
\node[anchor=south west, inner sep=2, font=\fontsize{8}{8}\selectfont] at (4.4000,21.3300) {0.3};
\fill[green!49!white] (4.3000,21.7250) rectangle (5.0500,22.1200);
\node[anchor=south west, inner sep=2, font=\fontsize{8}{8}\selectfont] at (4.4000,21.7250) {4.6};
\fill[green!24!white] (4.3000,22.1200) rectangle (5.0500,22.5150);
\node[anchor=south west, inner sep=2, font=\fontsize{8}{8}\selectfont] at (4.4000,22.1200) {2.2};
\fill[green!2!white] (4.3000,22.5150) rectangle (5.0500,22.9100);
\node[anchor=south west, inner sep=2, font=\fontsize{8}{8}\selectfont] at (4.4000,22.5150) {0.2};
\fill[green!40!white] (4.3000,22.9100) rectangle (5.0500,23.3050);
\node[anchor=south west, inner sep=2, font=\fontsize{8}{8}\selectfont] at (4.4000,22.9100) {3.7};
\fill[green!44!white] (4.3000,23.3050) rectangle (5.0500,23.7000);
\node[anchor=south west, inner sep=2, font=\fontsize{8}{8}\selectfont] at (4.4000,23.3050) {4.1};
\fill[green!39!white] (4.3000,23.7000) rectangle (5.0500,24.0950);
\node[anchor=south west, inner sep=2, font=\fontsize{8}{8}\selectfont] at (4.4000,23.7000) {3.6};
\node[anchor=south west, inner sep=2, rotate=20, yshift=0cm, xshift=0cm ,font=\fontsize{8}{8}\selectfont] at (4.3000,24.0950) {\textbf{\textcolor{red}{arXiv-T2V}}};
% +++++++++++++++++++++++++++++++++++++++++++++++++++++++++++++++

\end{tikzpicture}
}
\caption{Heat map of the most frequent aesthetic attributes over 4 datasets. The attributes are classified by different categories with colours, such as \colorbox{xy-axis}{x and y axes}, \colorbox{data-appearance}{data appearance}, \colorbox{annotation}{annotation}, \colorbox{title-subtitle}{title and subtitle}, \colorbox{legend}{legend}, \colorbox{grid}{grid}, \colorbox{figure-format}{figure format}, and \colorbox{others}{others}.} 
\label{fig:analysis2-attributes-full}
\end{figure}

\end{document}